\documentclass{article} %
\PassOptionsToPackage{numbers,compress}{natbib}
\usepackage[preprint]{preprint}
\usepackage{times}

\usepackage{amsmath,amsfonts,bm}

\def\eqref#1{equation~\ref{#1}}

\def\1{\bm{1}}

\DeclareMathAlphabet{\mathsfit}{\encodingdefault}{\sfdefault}{m}{sl}
\SetMathAlphabet{\mathsfit}{bold}{\encodingdefault}{\sfdefault}{bx}{n}

\usepackage{hyperref}
\usepackage{url}
\usepackage{fontawesome5}   %
\usepackage{amsmath,amssymb}
\usepackage{booktabs}
\usepackage[table]{xcolor}
\usepackage{arydshln}          %
\definecolor{oursblue}{HTML}{E4EEFA}   %
\definecolor{oursorange}{HTML}{FBEEE0}
\definecolor{linkblue}{HTML}{536D97}   %
\definecolor{divpos}{HTML}{E34948}\definecolor{divneg}{HTML}{2A78D6}
\usepackage{graphicx}
\usepackage{multirow}
\usepackage{algorithm}
\usepackage{enumitem}
\usepackage{algpseudocode}
\usepackage{tikz}
\usetikzlibrary{arrows.meta,positioning,fit}

\title{How Long, Not How Close: A Learned Temporal\\Metric for Planning in Latent World Models}

\author{Lama Moukheiber \qquad Haotian Xue \qquad Yongxin Chen \\[4pt]
\raisebox{-0.12\height}{\includegraphics[height=0.85em]{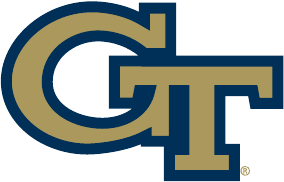}}~Georgia Institute of Technology \\[2pt]
{\hypersetup{pdfborder={0 0 0}}\small
\href{https://lamawmouk.github.io/tempo/}{\faGlobe~\textcolor{linkblue}{Project Page}} \qquad
\href{https://github.com/lamawmouk/tempo}{\faGithub~\textcolor{linkblue}{GitHub}}}
}

\makeatletter\renewcommand{\eqref}[1]{(\ref{#1})}\makeatother

\begin{document}

\maketitle

\begin{abstract}
Latent world models plan by rolling a frozen predictor forward under candidate action sequences and ranking the candidates by the
latent distance between their imagined end state and the goal. However, this ranking breaks down when the goal lies several plans away, because the latent distance measures how closely an end state resembles the goal rather than how
far it remains from reaching it. To address this, we propose \textbf{\textit{TEMPO}}, a temporal-distance planning objective that leaves the world model untouched, learns only from the recorded trajectories already used to train it, and adds negligible cost to the planner's search. \textit{TEMPO} learns a small map of the frozen latent in which
the distance between two states of an episode reflects the number of environment steps between them, and blends this distance into
the planner's cost. It requires no rewards, policies or success labels and, being a cost rather than a model, applies to frozen world models with one latent vector per state that plan by a latent distance. We evaluate \textit{TEMPO} on eleven simulated environments (e.g., maze
navigation, tabletop pushing, robotic arm control and three-dimensional manipulation) with the
LeWM and PLDM planners. With a small MLP that adds at most $0.3\%$ to a plan's arithmetic, \textit{TEMPO} improves
both planners at every goal distance, including the one-plan setting of their evaluations, raises LeWM from $36\%$ to
$99\%$ on TwoRoom three plans from the goal, and remains competitive on a broad range of 2D and 3D navigation, reaching and manipulation tasks.
\end{abstract}

\section{Introduction}
\label{sec:intro}

\begin{figure}[h!]
\centering
\includegraphics[width=0.85\textwidth]{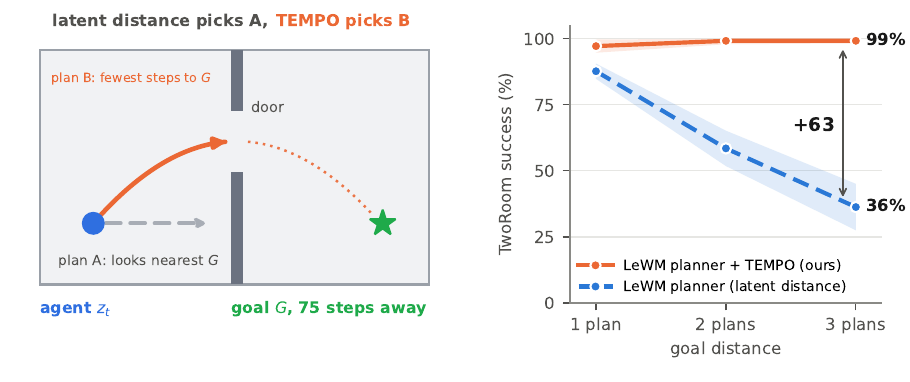}
\caption{\textbf{How long, not how close.} Left: three plans from the goal, the latent distance prefers the plan whose end state looks
most like the goal frame, toward the wall, whereas a distance that counts steps prefers the plan through the door. Right: TwoRoom success as the goal moves
from one to three plans away.}
\label{fig:teaser}
\end{figure}

A long-standing goal of learning-based control is an agent that plans directly from pixels, without hand-engineered state or a
task-specific policy. World models offer a path toward this goal. By predicting the consequences of its actions, an agent can search over them in imagination
and act on the best sequence it finds. Joint-embedding predictive architectures (JEPA) have become the practical
recipe for learning such models from video. An encoder maps each frame to a compact latent, a predictor rolls the latent forward
under actions, and both are trained to predict the embedding of the next frame rather than to reconstruct pixels. Planning then
amounts to model-predictive control in the latent space, which rolls the predictor out under candidate action sequences, keeps the one
whose imagined end state is closest to the encoded goal, executes it, and replans from the next observation. Recent planners of this
kind, such as PLDM and LeWM \citep{sobal2026learning,maes2026leworldmodel}, keep the model frozen after training, replan in closed loop and learn
no low-level policy. On the benchmarks these models are trained for, a single plan suffices to reach the goal, and LeWM's planner
succeeds on $94\%$ of Push-T windows whose goal lies 25 steps ahead, $88\%$ on TwoRoom and $95\%$ on Reacher.

Yet the goals that matter in practice, such as reaching the far side of a room through a door, pushing an object to a distant
target pose or placing an object in a sink, which takes about twelve plans in RoboCasa's demonstrations \citep{nasiriany2024robocasa}, lie well
beyond a single plan. The evaluations of these planners place the goal one plan away, whereas a task places it wherever
the task ends. When the
goal lies several plans away, no single action sequence the planner can score reaches it. Three plans away, the same planner collapses to $22\%$ on Push-T and $36\%$ on
TwoRoom, and every latent world model we test degrades in the same way (Figure~\ref{fig:teaser}). Such failures are usually attributed to compounding rollout error and a search space that grows with the horizon
\citep{zhang2026hierarchical}, both of which afflict planners that imagine the entire horizon at once. The planners we study instead imagine
a single plan, execute it and replan, reaching a distant goal through a sequence of short plans, and what breaks this sequence is the
cost itself, since every short plan ends far from a goal 75 steps out and ranking end states by their resemblance to the goal frame
reveals nothing about which plan leads there. The predictor is trained to
be locally accurate, whereas the planner reads latent distance as how much of the task remains, and the two agree within one plan of
the goal but diverge several plans away. Existing remedies each sacrifice part of what makes the frozen planner attractive. Retraining
or fine-tuning the world model with a temporal or reachability objective \citep{li2026predictive,bai2026temporal} changes the model,
which is not possible when it is frozen, and hierarchical planners
\citep{zhang2026hierarchical,cheng2026sage,caselli2026mind,hafner2022deep} insert a second decision process between the goal and the
planner and search over subgoals or macro-actions.

We introduce \textbf{\textit{TEMPO}} (\textit{\textbf{TEM}poral-distance \textbf{P}lanning \textbf{O}bjective}), a learned temporal distance for the planner's cost that leaves the world model untouched, draws only on the recorded trajectories already used to train it, and adds negligible cost to the planner's search. A small map $F_\phi$ embeds the frozen latent so that the distance between two states of a recorded
trajectory reflects the number of environment steps between them. It is trained by regression on pairs of frames from those trajectories, requires nothing beyond their frame indices, the time step at which each frame was recorded, and continues to learn online from the episodes the planner itself collects. Blended with the latent
distance rather than substituted for it, $F_\phi$ improves planning at every goal distance, and on the standard tasks most when the goal
is far, raising LeWM from $36$ to $99\%$ on TwoRoom and from $90$ to $98\%$ on Reacher three plans from the goal and from $22$ to $28\%$
on Push-T. The two distances are complementary. The latent distance captures the fine positioning that manipulation requires, while
$F_\phi$ measures \emph{how far from the goal in time} a state lies, and combining the two yields the strongest planner overall.

We evaluate \textit{TEMPO} on eleven environments spanning maze navigation, tabletop pushing, robotic arm control and
three-dimensional manipulation, on top of the LeWM and PLDM planners and on identical windows from one to
three plans away from the goal. Our experiments reveal the following findings:
\begin{itemize}\itemsep2pt
  \item \textit{TEMPO} enables a frozen short-horizon planner to reach goals several plans away. Three plans from the goal it improves
    the LeWM planner by $16$ points on average across our standard tasks, and LeWM with \textit{TEMPO} outperforms the PLDM planner on every standard task;
    it takes LeWM from $36$ to $99\%$ on TwoRoom and roughly quadruples it on cross-room TwoRoom (Section~\ref{sec:results}).
  \item \textit{TEMPO} improves both planners at every goal distance, including the one-plan setting of their evaluations (Section~\ref{sec:results}).
  \item \textit{TEMPO} can be trained across environment variations within a task family (e.g., different wall layouts for navigation
    or different block shapes for pushing) and, without retraining, achieves higher success than the planner under layout shifts (unseen wall layouts and door positions) and shape shifts
    (unseen block shapes), tripling LeWM's success with a moved door three plans from the goal (Section~\ref{sec:gen}).
  \item \textit{TEMPO} is computationally lightweight. It requires the same number of predictor rollouts as the original planner,
    introduces no second model or level of search, and increases the cost of each plan by at most $0.3\%$ without task-specific
    tuning (Appendix~\ref{app:cem}).
  \item \textit{TEMPO} ranks plans by whether they will reach the goal. Three plans from the goal, the latent distance tracks where a plan ends,
    while \textit{TEMPO}'s cost ranks the same candidate plans by their eventual success, which is what the planner needs (Appendix~\ref{sec:geometry}).
\end{itemize}

\section{Related work}
\label{sec:related}

\paragraph{JEPA world models.}
World models learn the environment's dynamics from data so that an agent can reason about future states in imagination \citep{ha2018recurrent}. Generative world models simulate future observations in pixel space
\citep{hafner2019learning,hafner2019dream,hafner2025mastering,micheli2022transformers,alonso2024diffusion}. Joint-embedding predictive architectures (JEPA) instead predict in a compact latent space
\citep{lecun2022path}, first as self-supervised representation learners (I-JEPA, V-JEPA \citep{assran2023self,bardes2024revisiting}) and,
trained on reward-free offline trajectories, as action-conditioned world models that plan zero-shot for goals given as images. PLDM and LeWM encode each frame as a single latent
vector. PLDM learns its encoder and predictor jointly from pixels with a VICReg-style regulariser
\citep{sobal2026learning,bardes2021vicreg}, and LeWM trains a ViT encoder end to end with LeJEPA's single Gaussian regulariser and plans
with CEM over its 192-dimensional class token \citep{maes2026leworldmodel,balestriero2025lejepa}. DINO-WM instead predicts a grid of frozen
DINOv2 patch features per frame and so cannot collapse \citep{zhou2024dino,oquab2023dinov2}, but its pretrained encoder and many
tokens per frame make it larger and much slower to train and plan than models with a single latent vector \citep{maes2026leworldmodel}, and its cost also reads the agent's proprioceptive state, so we focus here on single-latent world models.

\paragraph{Planning with latent dynamics.}
Dreamer and TD-MPC learn a policy or value in imagination and need rewards \citep{hafner2019dream,hansen2022temporal,hansen2024td}, while offline goal-conditioned RL learns a policy or value directly \citep{ghosh2019learning,kostrikov2021offline,park2023hiql}. Reward-free JEPA
planners instead query the frozen dynamics by sampling-based trajectory optimisation \citep{rubinstein2004cross,pinneri2021sample},
executing the candidate whose imagined end state is closest to the encoded goal \citep{zhou2024dino,sobal2026learning,maes2026leworldmodel}. Their published evaluations
\citep{maes2026leworldmodel,zhou2024dino,park2025ogbench} stop one plan from the goal, whereas we evaluate several plans out,
with every arm on identical windows, and add layout, shape and visual shifts at test time \citep{wang2026adajepa}.

\paragraph{Long-horizon planning and latent distances.}
Learned world models are known to struggle over long horizons, as rollout errors compound and the search space grows with the horizon
\citep{zhang2026hierarchical}. Hierarchical planners address this with coarse models or subgoal generators
\citep{zhang2026hierarchical,cheng2026sage,caselli2026mind,pertsch2020long,nasiriany2019planning,hafner2022deep,park2023hiql}, at the
price of a second model or level of search, while a closed-loop planner that replans after each short plan avoids both issues yet still
fails several plans out, because its distance ranks resemblance to the goal rather than distance from it. Another line of work therefore
makes latent distances track progress. RC-aux \citep{li2026predictive} adds a reachability-correction objective to LeWM, TD-JEPA
\citep{bai2026temporal} retrains the encoder so that its latent distance reflects temporal progress, Hilbert representations
\citep{park2024foundation} learn a space whose distance counts steps, and AdaJEPA \citep{wang2026adajepa} adapts the world model inside
MPC, but all of these change the representation or the predictor and so require a model that can be retrained. We instead propose to
leave the world model frozen and learn a metric embedding of the frozen latent whose distance counts steps, updated from the planner's
own episodes and blended into the planner's cost.

\section{Method: \textit{TEMPO}}
\label{sec:method}
\label{sec:prelim}

In this section we introduce \textit{TEMPO}, a temporal-distance planning objective for frozen world models with one latent vector per state that plan by a latent distance to the goal, such as LeWM and PLDM. We first state
the problem setting, namely the offline reward-free data, the latent world model and the planner that every method in this paper shares.
We then describe the learned temporal distance $F_\phi$ and its training objective, and finally how it enters the planner's cost.

\subsection{Problem setting}
\label{sec:setting}

\paragraph{Offline reward-free data.}
We consider goal-conditioned control from pixels. The only training data is an offline set of trajectories
\begin{equation}
  \mathcal D = \{\tau^{(n)}\}_{n=1}^{N}, \qquad \tau = (o_{1:T},\, a_{1:T-1}),
  \label{eq:data}
\end{equation}
where $o_t$ is an image and $a_t$ the action taken between $o_t$ and $o_{t+1}$, with no rewards, task labels or success
signals, and the behaviour policies that produced the trajectories need not be optimal. A task is specified at test time by a goal
image $o_G$.

\paragraph{Latent world model.}

\begin{figure}[t]
\centering
\includegraphics[width=\textwidth]{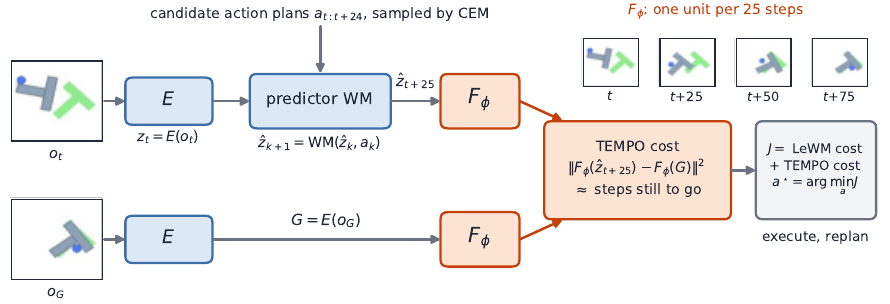}
\caption{\textbf{\textit{TEMPO} in the planner.} LeWM's frozen encoder $E$ and predictor $\mathrm{WM}$ (blue) are unchanged, and nothing is decoded. The planner rolls $\mathrm{WM}$ out under each CEM candidate plan and executes the one that minimises $J$. \textit{TEMPO} adds the map $F_\phi$ (orange) and scores each candidate plan by $\lVert F_\phi(\hat z_{t+25}) - F_\phi(G)\rVert^2$, the
steps its end state still is from the goal, added to LeWM's cost. Top right: $F_\phi$ places the frames of one episode one unit apart
per 25 steps.}
\label{fig:pipeline}
\end{figure}

The world models we study are joint-embedding predictive architectures (Figure~\ref{fig:pipeline}), in which an encoder maps an observation to a
latent, $z_t = E(o_t)$, and an action-conditioned predictor advances it, $\hat z_{t+1} = \mathrm{WM}(z_t, a_t)$. Both are trained on
$\mathcal D$ to predict in embedding space rather than to reconstruct pixels, with the one-step latent prediction error
$\lVert \mathrm{WM}(z_t, a_t) - E(o_{t+1})\rVert^2$ plus, when the encoder is trained, a regulariser that keeps the embedding from collapsing. LeWM and PLDM differ only in that regulariser, and both are single-latent world models that compress a frame into one 192-d latent vector, with actions in blocks of
5 environment steps. Neither has a decoder, a reward model or a policy, and both are frozen once trained.

\paragraph{Planning by latent distance.}
With $G = E(o_G)$, model-predictive control rolls the predictor forward under a candidate action sequence $a = a_{t:t+H-1}$,
$\hat z_{t+k} = \mathrm{WM}(\hat z_{t+k-1}, a_{t+k-1})$ with $\hat z_t = z_t$, and scores the imagined trajectory by a terminal, a
running and an action term,
\begin{equation}
  J(a) = w_T\,\lVert \hat z_{t+H} - G\rVert^2 + w_r \sum_{k=1}^{H-1} \lVert \hat z_{t+k} - G\rVert^2 + w_a \sum_{k} \lVert a_{t+k}\rVert^2,
  \qquad a^\star = \arg\min_a J(a).
  \label{eq:mpc-general}
\end{equation}
The published planners are the terminal-only instance, $w_T = 1$, $w_r = w_a = 0$,
\begin{equation}
  a^\star = \arg\min_{a} \;\lVert \hat z_{t+H}(z_t, a) - G \rVert^2,
  \label{eq:flat}
\end{equation}
solved by the cross-entropy method (300 sequences, 30 iterations, top 30, and $H = 25$ steps as 5 blocks of 5) and executed in full before
the state is re-encoded and Eq.~\eqref{eq:flat} re-solved. LeWM and PLDM score the learned latent alone. This is the flat planner of every table, and the object \textit{TEMPO} changes.

\subsection{Learning the temporal distance}
\label{sec:fmap}

$F_\phi$ is a small MLP (two hidden layers of 256 units, SiLU) from the 192-d frozen latent $z$ to a 64-d embedding, trained so that distance in the
embedding counts environment steps. On pairs of frames of the same episode $(z_t, z_{t+k})$,
\begin{equation}
  \begin{aligned}
  \mathcal L_F = {}& \mathrm{smooth}\text{-}L_1\!\Big(\lVert F_\phi(z_t) - F_\phi(z_{t+k})\rVert,\ \tfrac{k}{\kappa}\Big)_{k \le 75} \\
  &+ \tfrac12\,\max\!\big(0,\ 3 - \lVert F_\phi(z_t) - F_\phi(z_{t+d})\rVert\big)_{d > 75},
  \quad \kappa = 25,\ 3 = 75/\kappa,
  \end{aligned}
  \label{eq:fmap}
\end{equation}
so that a 25-step displacement has unit length. The regression targets $k/\kappa$ therefore run up to $75/\kappa = 3$, the distance of
the farthest goal we evaluate, three plans away. Pairs more than 75 steps apart are not regressed to an exact distance, which would be
noisy that far out. The hinge only pushes them at least $3$ units apart, the largest regression target, so that a state more than three
plans from the goal never appears closer to it than a state within three plans. The metric is thus calibrated up to three plans and
only orders states beyond that, and the factor $\tfrac12$ gives the hinge half the weight of the regression term. The loss uses only the time steps of the recorded trajectories, with no reward,
environment state, action or success label. We call $k$ the \emph{trajectory temporal distance}: the number of steps a recorded trajectory took between two of its states. It is the training target; the quantity planning ultimately needs, the fewest steps from a state to the goal, is never observed, and Tables~\ref{tab:actionrank} and~\ref{tab:rank-tempo} show that the learned distance ranks plans by their eventual success. $F_\phi$ is first trained offline on windows of these trajectories (Adam $10^{-3}$, batch 256,
gradient clipping 10, 3,000 steps on 20,000 windows) and then keeps learning online. For eight rounds of 50 training windows the planner
collects its own episodes in 50 parallel environments and $F_\phi$ takes one gradient step on every stored same-episode pair after each
decision batch. The offline windows come from the trajectories the world model is trained on and the online episodes are the planner's own, so
\textit{TEMPO} needs no data or labels beyond what the world model already uses.

\subsection{Planning with \textit{TEMPO}}
\label{sec:planning}

The planner of Eq.~\eqref{eq:flat} keeps its world model, CEM, horizon and cadence and minimises
\begin{equation}
  J(a) = (1 - w)\,\lVert \hat z_{t+H}(z_t, a) - G\rVert^2 \;+\; w\,\bar d_H^{\,2}\,\lVert F_\phi(\hat z_{t+H}(z_t, a)) - F_\phi(G)\rVert^2,
  \qquad w = 0.5,
  \label{eq:cost}
\end{equation}
where $\bar d_H$ is the median 25-step latent displacement of the training data, so that the two terms are in latent units and the blend weights them equally. The raw latent keeps the fine position that the temporal coordinate does not resolve, and the temporal
term orders plans by how far their end state still is from the goal in time. We use this blend rather than $F_\phi$ alone in every
experiment because it is better on average for both LeWM and PLDM at every goal distance on the five standard environments
(Appendix Table~\ref{tab:ablation}). Every $H = 25$ environment steps, with $z_t = E(o_t)$
and $G = E(o_G)$ from the frozen encoder,
\begin{align}
  \hat z_{t+k} &= \mathrm{WM}(\hat z_{t+k-1}, a_{t+k-1}), \qquad \hat z_t = z_t, \qquad k = 1, \dots, H, \label{eq:app-roll}\\
  a^\star_{t:t+H} &= \arg\min_a J(a_{t:t+H}) \quad \text{by CEM with the cost of Eq.~\eqref{eq:cost}; execute all } H \text{ actions and re-encode}, \label{eq:app-act}\\
  \phi &\leftarrow \phi - \eta\, \nabla_\phi \mathcal L_F \quad \text{(online rounds: one step per decision batch on the stored pairs).} \label{eq:app-update}
\end{align}
The flat planner is Eqs.~\eqref{eq:app-roll} and \eqref{eq:app-act} with $w = 0$, and \textit{TEMPO} adds the second term of
Eq.~\eqref{eq:cost} and the update of Eq.~\eqref{eq:app-update}. $F_\phi$ is evaluated in batch on the CEM's own predicted end states,
one small MLP forward pass per candidate against the $H$ predictor steps that produced them, so it adds well under $1\%$ of the compute
of a plan (Appendix~\ref{app:method}). Control is closed-loop at the plan level, as for every method. Evaluation freezes everything, with 5 fixed groups of 50 windows per world-model seed and three seeds, with defaults in Appendix~\ref{app:method}. Why a distance that counts steps orders candidate plans where raw latent similarity does not is explained in Appendix~\ref{sec:why}, which also shows that $F_\phi$ is goal-agnostic, generalises to held-out episodes and does not collapse to the frame index.

\section{Experiments}
\label{sec:experiments}

Our experiments aim to answer the following questions: 1) Does \textit{TEMPO} improve a frozen flat planner
on goals several plans away? 2) Does it improve both frozen world models, LeWM and PLDM (Section~\ref{sec:results})? 3) Does it transfer,
without retraining, to environment configurations unseen in training (Section~\ref{sec:gen})? 4) Which part of
\textit{TEMPO}'s cost drives the gain, and why does the latent-distance cost fail far from the goal
(Section~\ref{sec:ablation})?

\subsection{Environments and tasks}
\label{sec:setup}

We use eleven environments spanning navigation, reaching and manipulation in two and three dimensions
\citep{maes2026leworldmodel,zhou2024dino,park2025ogbench,fu2020d4rl}, all run on stable-worldmodel \citep{maes2026stable}. They range from 2D point navigation (TwoRoom, Wall, WallRandom, PointMaze medium and large) through two-joint arm reaching
(Reacher) and contact-rich tabletop pushing (Push-T, PushObj) to 3D manipulation with a robot arm (Cube, Scene) and a legged robot
(Ant-U-Maze). Every trial starts from a state of a recorded expert episode, and the agent must reach a goal given only as an image, the expert's frame $25$, $50$ or $75$ steps later, one to three plans away. The agent sees nothing but $224\times224$ RGB frames,
and success follows each environment's own criterion. Details on the environments and datasets can be found in Appendix~\ref{app:envs}.

\subsection{Baselines and evaluation protocol}
\label{sec:protocol}

\paragraph{Baselines.}
We evaluate \textit{TEMPO} on two JEPA world-model planners. Both learn from the offline trajectories alone, with no rewards or task labels,
and plan as published through the same CEM on the same windows, and \textit{TEMPO} is then added to each one's cost with
everything else unchanged, so that each pair isolates the effect of the cost.
\begin{enumerate}[label=\alph*),leftmargin=*,itemsep=1pt,topsep=2pt]
\item \textbf{LeWM} \citep{maes2026leworldmodel}: a recent JEPA world model that trains a ViT encoder and a latent predictor end to end from pixels with a prediction loss
and a single anti-collapse regulariser (SIGReg), and plans by minimising the latent distance to the encoded goal. The frozen LeWM model is
the instance on which every claim of this paper is made.
\item \textbf{PLDM} \citep{sobal2026learning}: a world model that learns its encoder and predictor jointly with a VICReg-style multi-term objective
(prediction, variance, covariance and temporal smoothness terms) and plans with the same latent-distance cost. We implement it with
PLDM's objective on LeWM's ViT encoder and a 192-d latent, as in LeWM's baselines.
\end{enumerate}
LeWM and PLDM are the current end-to-end JEPA latent-dynamics models learned from pixels, each with one compact latent per frame and $18$M
parameters, planning by the latent distance \textit{TEMPO} corrects. TwoRoom, Reacher, Push-T and Cube are LeWM's benchmark tasks with its released datasets, and one plan from the goal is LeWM's own
protocol (goal 25 steps ahead, budget 50), and on these tasks LeWM is the primary comparator. We do not include offline goal-conditioned
reinforcement-learning and imitation methods such as GCBC, IQL and IVL \citep{park2023hiql,park2025ogbench}, which learn a policy or value function from the data, whereas
\textit{TEMPO} is not a reinforcement-learning method.

\paragraph{Training.} Every world model is trained once on the released offline expert set for its environment, 1,789 to 20,000
reward-free trajectories (Table~\ref{tab:datasets}), with the 10-epoch recipe of \citet{maes2026leworldmodel}, three seeds per environment, and is then frozen, with hyperparameters fixed across environments (Appendices~\ref{app:method} and~\ref{app:baselines}). $F_\phi$ learns in two stages, first offline, on windows cut from the same recorded trajectories, and then online, from the episodes the planner itself generates
during eight online rounds (Section~\ref{sec:fmap}). Its architecture, loss and schedule are identical everywhere.

\paragraph{Planning.} Every planner optimises a plan of $H = 5$ action blocks (25 environment steps) with the cross-entropy method (CEM) \citep{maes2026leworldmodel,zhou2024dino}, with LeWM's settings
(300 candidates, 30 iterations and 30 elites, Appendix~\ref{app:cem}). The whole plan is executed, the state re-encoded and the plan
re-solved until the episode ends. We measure goal distance in plans, so a goal $k$ plans away is the frame $25k$ environment steps
later in the same recorded trajectory, so beyond one plan no single plan reaches it and the planner must chain at least $k$ of them, and the planner is the same at every goal distance, only the goal changes.
At that range the latent distance acts like a straight-line cost that can stall against obstacles, whereas $F_\phi$, learned from
trajectories, counts steps along feasible paths, so each short plan makes progress along the true route. The only difference between a planner and its \textit{TEMPO} variant is the cost the CEM minimises,
Eq.~\eqref{eq:flat} for the planner and Eq.~\eqref{eq:cost} for \textit{TEMPO}.

\paragraph{Evaluation.} An evaluation window is a slice of a recorded expert episode, in which the environment is reset to frame $s$ and the
planner is shown the expert's frame at $s + \text{offset}$ as the goal. Offset 25 is
the published setting, one plan away, while offsets 50 and 75 ask for states two and three plans away, beyond anything the predictor is
rolled out to. Success is mean $\pm$ std error over 15 evaluations (5 groups $\times$ 3 seeds) of 250 identical windows per seed.
In tests outside the training distribution, $F_\phi$ is
trained in the training environment and kept fixed during evaluation, so any gain under the shift comes from the temporal distance.

\subsection{TEMPO: learned temporal distance for planning with frozen world models}
\label{sec:results}

\paragraph{The flat planners collapse with distance.} Tables~\ref{tab:std75} and~\ref{tab:complex} report planning success for every
planner one, two and three plans from the goal, on identical windows, where each ``$+\,$\textit{TEMPO}'' row keeps the planner above
it and changes only the cost it minimises. Both flat planners succeed one plan from the goal but degrade as it moves
away, most sharply on TwoRoom and Push-T. The collapse is
therefore a property of the objective these planners share, a latent distance to a far goal, rather than of any one world model.

\paragraph{\textit{TEMPO} closes the gap.} \textit{TEMPO} consistently improves both planners at every horizon, and on the standard
tasks its gain is largest three plans from the goal, where the flat planners collapse ($\Delta$ column of Table~\ref{tab:std75}). Remarkably, three plans from the goal it takes LeWM on TwoRoom and Reacher to near-perfect success, and on the more complex tasks of Table~\ref{tab:complex} it more than doubles PointMaze medium
one plan out. On the remaining tasks, \textit{TEMPO} remains competitive. Additionally, the same pattern holds for PLDM, and LeWM with \textit{TEMPO} consistently outperforms the published PLDM planner on every
core environment at every goal distance. These gains come at negligible cost, since both variants roll the same candidates through the predictor,
and \textit{TEMPO} adds one small MLP evaluation per candidate, at most $0.3\%$ of a plan's arithmetic (Appendix~\ref{app:cem}).

\begin{figure}[t]
\centering
\includegraphics[width=\textwidth]{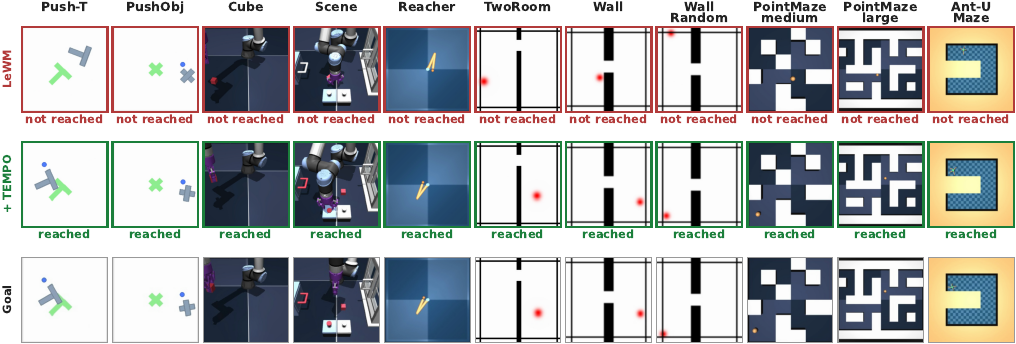}
\caption{\textbf{How long, not how close.} Final frames of one episode per environment, where the goal is the expert's own frame 75 steps later (Wall and WallRandom: 40 and 25). \textit{Reached}: the episode met the environment's success test within its step budget; \textit{not reached}: it did not.}
\label{fig:rollouts}
\end{figure}

\paragraph{Cross-room TwoRoom.} On PLDM's own Two-Rooms evaluation \citep{sobal2026learning}, the goal always lies in the other room, so
every trial must find the door. Each group uses a fixed set of 50 start and goal pairs in different rooms, and the one-, two- and three-plan
rows use a fixed budget of 50, 100 and 150 steps. Using the same world models and the same $F_\phi$ without refitting, \textit{TEMPO} roughly
quadruples LeWM's success at every budget (Table~\ref{tab:complex}). Figure~\ref{fig:rollouts} shows one episode per environment.

\begin{table}[t]
\centering\footnotesize
\caption{\textbf{Success rates (\%) on five pixel-based goal-conditioned control tasks, one to three plans from the goal.} Bold marks the better of each pair. $\Delta$ is the gain of \textit{TEMPO} over the same planner, averaged over environments. Every world model plans from pixels alone.}
\label{tab:std75}
\setlength{\tabcolsep}{4pt}
\footnotesize %
\resizebox{\ifdim\width>\linewidth\linewidth\else\width\fi}{!}{\scalebox{0.87}{%
\begin{tabular}{lccccccc}
\toprule
\textbf{Planner} & \textbf{TwoRoom} & \textbf{Reacher} & \textbf{Push-T} & \textbf{Cube} & \textbf{Scene} & \textbf{Avg.} & \textbf{$\Delta$} \\
\midrule
\rowcolor{gray!12}\multicolumn{8}{l}{\textbf{LeWM}} \\
one plan & 87.7{\scriptsize$\pm$0.8} & 94.7{\scriptsize$\pm$0.7} & 93.6{\scriptsize$\pm$1.0} & 74.8{\scriptsize$\pm$2.4} & 69.6{\scriptsize$\pm$1.1} & 84.1 & – \\
\rowcolor{oursblue}\quad \textbf{+ TEMPO (ours)} & \textbf{97.2{\scriptsize$\pm$0.6}} & \textbf{95.5{\scriptsize$\pm$0.7}} & \textbf{95.1{\scriptsize$\pm$0.6}} & \textbf{77.1{\scriptsize$\pm$1.7}} & \textbf{71.1{\scriptsize$\pm$1.1}} & \textbf{87.2} & +3.1 \\
\addlinespace[1.5pt]
two plans & 58.5{\scriptsize$\pm$1.7} & 95.7{\scriptsize$\pm$0.6} & 49.7{\scriptsize$\pm$1.3} & 57.9{\scriptsize$\pm$2.4} & 56.4{\scriptsize$\pm$1.2} & 63.7 & – \\
\rowcolor{oursblue}\quad \textbf{+ TEMPO (ours)} & \textbf{99.2{\scriptsize$\pm$0.3}} & \textbf{96.8{\scriptsize$\pm$0.5}} & \textbf{55.7{\scriptsize$\pm$1.2}} & \textbf{58.7{\scriptsize$\pm$2.0}} & \textbf{57.9{\scriptsize$\pm$1.3}} & \textbf{73.7} & +10.0 \\
\addlinespace[1.5pt]
three plans & 36.3{\scriptsize$\pm$2.3} & 89.7{\scriptsize$\pm$0.4} & 22.0{\scriptsize$\pm$0.6} & 68.7{\scriptsize$\pm$1.6} & 48.7{\scriptsize$\pm$1.3} & 53.1 & – \\
\rowcolor{oursblue}\quad \textbf{+ TEMPO (ours)} & \textbf{99.2{\scriptsize$\pm$0.3}} & \textbf{98.4{\scriptsize$\pm$0.5}} & \textbf{28.1{\scriptsize$\pm$2.0}} & \textbf{70.4{\scriptsize$\pm$1.2}} & \textbf{50.5{\scriptsize$\pm$1.3}} & \textbf{69.3} & +16.2 \\
\addlinespace[4pt]
\midrule
\rowcolor{gray!12}\multicolumn{8}{l}{\textbf{PLDM}} \\
one plan & 96.8{\scriptsize$\pm$0.4} & 93.6{\scriptsize$\pm$1.0} & 76.1{\scriptsize$\pm$1.8} & 60.3{\scriptsize$\pm$2.3} & 56.1{\scriptsize$\pm$1.8} & 76.6 & – \\
\rowcolor{oursblue}\quad \textbf{+ TEMPO (ours)} & \textbf{98.9{\scriptsize$\pm$0.3}} & \textbf{94.5{\scriptsize$\pm$1.0}} & \textbf{90.1{\scriptsize$\pm$1.2}} & \textbf{68.7{\scriptsize$\pm$1.4}} & \textbf{58.4{\scriptsize$\pm$1.3}} & \textbf{82.1} & +5.5 \\
\addlinespace[1.5pt]
two plans & 82.0{\scriptsize$\pm$2.0} & 93.6{\scriptsize$\pm$0.6} & 31.9{\scriptsize$\pm$1.5} & 48.5{\scriptsize$\pm$1.7} & 43.6{\scriptsize$\pm$0.9} & 59.9 & – \\
\rowcolor{oursblue}\quad \textbf{+ TEMPO (ours)} & \textbf{97.2{\scriptsize$\pm$0.5}} & \textbf{98.0{\scriptsize$\pm$0.4}} & \textbf{39.5{\scriptsize$\pm$1.8}} & \textbf{48.7{\scriptsize$\pm$1.8}} & \textbf{45.9{\scriptsize$\pm$1.2}} & \textbf{65.8} & +5.9 \\
\addlinespace[1.5pt]
three plans & 70.5{\scriptsize$\pm$1.4} & 88.0{\scriptsize$\pm$0.5} & 11.6{\scriptsize$\pm$1.3} & 65.6{\scriptsize$\pm$1.7} & 38.7{\scriptsize$\pm$1.7} & 54.9 & – \\
\rowcolor{oursblue}\quad \textbf{+ TEMPO (ours)} & \textbf{95.7{\scriptsize$\pm$1.1}} & \textbf{97.7{\scriptsize$\pm$0.5}} & \textbf{17.1{\scriptsize$\pm$1.7}} & \textbf{66.7{\scriptsize$\pm$2.0}} & \textbf{41.1{\scriptsize$\pm$1.5}} & \textbf{63.7} & +8.8 \\
\bottomrule
\end{tabular}}}
\end{table}

\begin{table}[t]
\centering\footnotesize
\caption{\textbf{Success rates (\%) on more complex goal-conditioned control tasks, one to three plans from the goal.} Reporting, bold and $\Delta$ as in Table~\ref{tab:std75}. $^\dagger$Wall episodes end at 50 steps, so its two-plan cell uses goal 40 and it has no three-plan cell.}
\label{tab:complex}
\setlength{\tabcolsep}{4pt}
\footnotesize %
\resizebox{\ifdim\width>\linewidth\linewidth\else\width\fi}{!}{\scalebox{0.87}{%
\begin{tabular}{lccccccc}
\toprule
\textbf{Planner} & \textbf{Wall$^\dagger$} & \textbf{\shortstack{TwoRoom\\cross-room}} & \textbf{PointMaze M} & \textbf{PointMaze L} & \textbf{Ant-U} & \textbf{Avg.} & \textbf{$\Delta$} \\
\midrule
\rowcolor{gray!12}\multicolumn{8}{l}{\textbf{LeWM}} \\
one plan & 67.6{\scriptsize$\pm$1.4} & 12.0{\scriptsize$\pm$1.0} & 32.1{\scriptsize$\pm$1.4} & 28.0{\scriptsize$\pm$2.0} & 19.7{\scriptsize$\pm$1.0} & 31.9 & – \\
\rowcolor{oursblue}\quad \textbf{+ TEMPO (ours)} & \textbf{75.7{\scriptsize$\pm$2.2}} & \textbf{46.5{\scriptsize$\pm$1.2}} & \textbf{75.7{\scriptsize$\pm$1.7}} & \textbf{54.8{\scriptsize$\pm$1.6}} & \textbf{21.9{\scriptsize$\pm$1.1}} & \textbf{54.9} & +23.0 \\
\addlinespace[1.5pt]
two plans & 57.7{\scriptsize$\pm$2.4} & 15.5{\scriptsize$\pm$1.0} & 17.6{\scriptsize$\pm$1.0} & 2.9{\scriptsize$\pm$0.6} & 25.5{\scriptsize$\pm$1.5} & 23.8 & – \\
\rowcolor{oursblue}\quad \textbf{+ TEMPO (ours)} & \textbf{65.9{\scriptsize$\pm$2.2}} & \textbf{69.6{\scriptsize$\pm$1.8}} & \textbf{41.1{\scriptsize$\pm$2.0}} & \textbf{21.3{\scriptsize$\pm$1.2}} & \textbf{29.6{\scriptsize$\pm$1.0}} & \textbf{45.5} & +21.7 \\
\addlinespace[1.5pt]
three plans & – & 17.1{\scriptsize$\pm$1.1} & 10.8{\scriptsize$\pm$0.8} & 0.3{\scriptsize$\pm$0.2} & 28.1{\scriptsize$\pm$1.6} & 14.1 & – \\
\rowcolor{oursblue}\quad \textbf{+ TEMPO (ours)} & – & \textbf{74.5{\scriptsize$\pm$1.7}} & \textbf{31.7{\scriptsize$\pm$1.0}} & \textbf{6.5{\scriptsize$\pm$1.0}} & \textbf{34.0{\scriptsize$\pm$2.1}} & \textbf{36.7} & +22.6 \\
\addlinespace[4pt]
\midrule
\rowcolor{gray!12}\multicolumn{8}{l}{\textbf{PLDM}} \\
one plan & 71.2{\scriptsize$\pm$1.9} & 23.9{\scriptsize$\pm$1.5} & 22.8{\scriptsize$\pm$2.1} & 32.3{\scriptsize$\pm$1.5} & 17.1{\scriptsize$\pm$1.5} & 33.5 & – \\
\rowcolor{oursblue}\quad \textbf{+ TEMPO (ours)} & \textbf{82.3{\scriptsize$\pm$1.7}} & \textbf{42.5{\scriptsize$\pm$1.5}} & \textbf{43.1{\scriptsize$\pm$1.8}} & \textbf{48.5{\scriptsize$\pm$2.9}} & \textbf{18.5{\scriptsize$\pm$1.7}} & \textbf{47.0} & +13.5 \\
\addlinespace[1.5pt]
two plans & 63.5{\scriptsize$\pm$2.4} & 35.2{\scriptsize$\pm$1.7} & 16.8{\scriptsize$\pm$1.0} & 6.5{\scriptsize$\pm$0.7} & 24.3{\scriptsize$\pm$1.1} & 29.3 & – \\
\rowcolor{oursblue}\quad \textbf{+ TEMPO (ours)} & \textbf{72.5{\scriptsize$\pm$1.4}} & \textbf{62.0{\scriptsize$\pm$1.6}} & \textbf{26.9{\scriptsize$\pm$2.1}} & \textbf{15.1{\scriptsize$\pm$1.8}} & \textbf{29.3{\scriptsize$\pm$1.7}} & \textbf{41.2} & +11.9 \\
\addlinespace[1.5pt]
three plans & – & 39.1{\scriptsize$\pm$1.8} & 8.7{\scriptsize$\pm$0.8} & 3.1{\scriptsize$\pm$0.8} & 26.8{\scriptsize$\pm$1.5} & 19.4 & – \\
\rowcolor{oursblue}\quad \textbf{+ TEMPO (ours)} & – & \textbf{65.1{\scriptsize$\pm$1.7}} & \textbf{16.7{\scriptsize$\pm$0.8}} & \textbf{7.1{\scriptsize$\pm$1.0}} & \textbf{29.3{\scriptsize$\pm$1.7}} & \textbf{29.5} & +10.1 \\
\bottomrule
\end{tabular}}}
\end{table}

\subsection{Towards generalisation to novel environment configurations}
\label{sec:gen}
\begin{figure}[!t]
\centering
\includegraphics[width=0.95\textwidth]{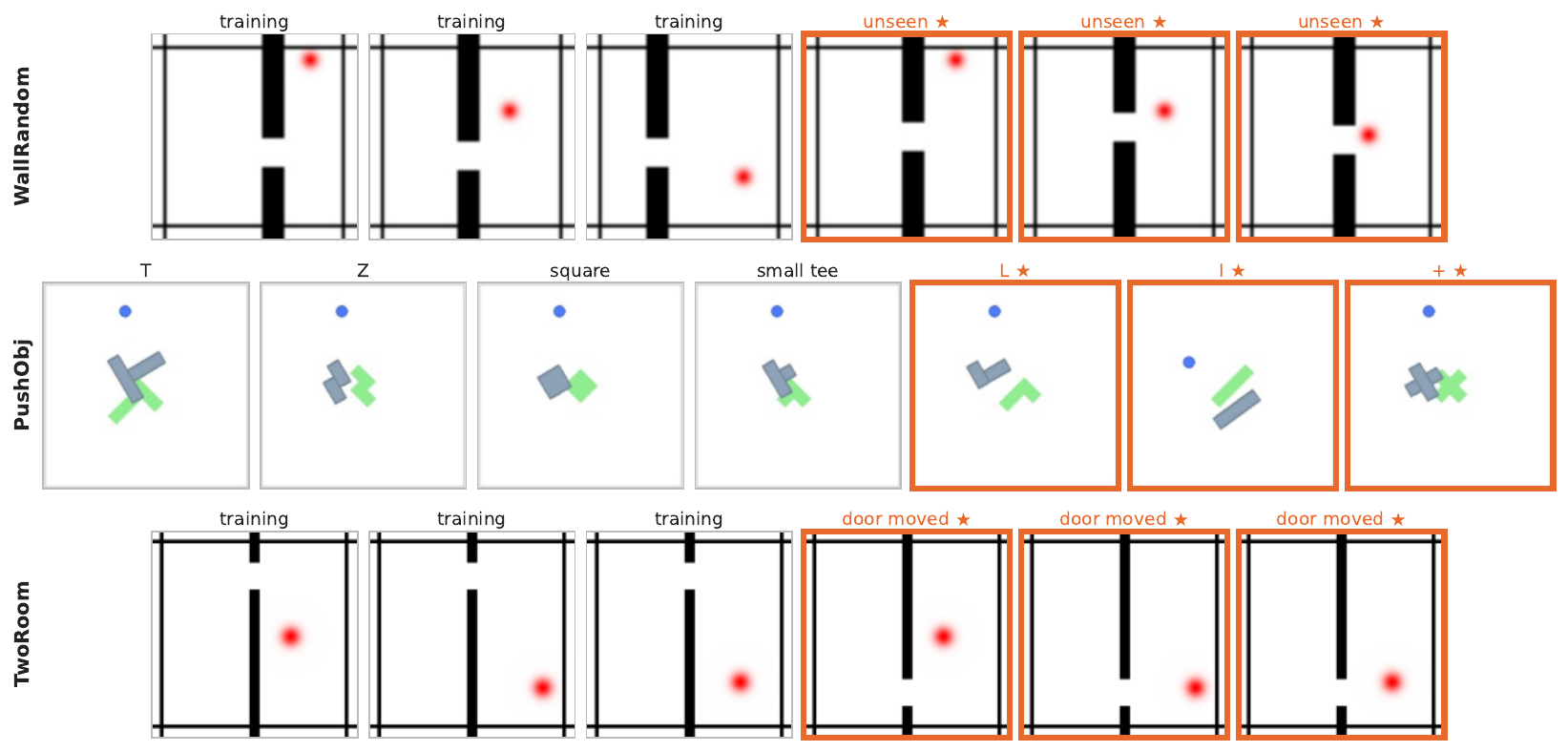}
\vspace{-10pt}
\caption{\textbf{Training and test setups for the three generalisation families of Table~\ref{tab:gen}.} Unseen configurations are
outlined in orange ($\star$). WallRandom holds out wall and door positions outside every training range, PushObj holds out the L, I
and + shapes, and TwoRoom moves the door to the opposite end of the wall.}
\label{fig:gen-setups}
\vspace{-6pt}
\end{figure}

Beyond new goals, we also evaluate \textit{TEMPO} on environments that differ from training, using three families with test configurations unseen in training, namely WallRandom \citep{zhou2024dino}, with unseen wall and door layouts (a layout shift)
and a cross-wall test that asks the planner to cross an unseen wall, PushObj \citep{zhou2024dino}, with unseen block shapes (a shape shift), and TwoRoom with
the door moved to a position never seen in training. The world model,
the planner and $F_\phi$ see only the training configurations (Figure~\ref{fig:gen-setups}). Notably, \textit{TEMPO}
substantially improves the LeWM planner on unseen wall layouts and on the cross-wall test. It also improves PLDM on PushObj's unseen shapes and remains competitive for LeWM, and on the moved door
it substantially improves both planners three plans out, where every planner's success is lowest (Table~\ref{tab:gen}). Beyond configurations, \textit{TEMPO} also keeps its advantage under test-time visual and dynamics shifts, improving the
planner under blur and colour jitter on TwoRoom and Push-T and with a larger agent or a heavier arm, and remaining competitive under
the remaining corruptions and camera changes (Appendix~\ref{app:shift}, Tables~\ref{tab:corruption}, \ref{tab:corruption-complex}
and~\ref{tab:env-shift}). We further extend the standard and more complex tasks to a mobile manipulator in RoboCasa kitchens
\citep{nasiriany2024robocasa}, covering the core skills of pick-and-place and navigation (Appendix~\ref{app:robocasa}).

\begin{table}[t]
\centering\small
\caption{\textbf{Generalisation to unseen configurations: a layout shift (WallRandom), a shape shift (PushObj) and a moved door (TwoRoom).} World models and TEMPO are trained on the training layouts, block shapes or door position; in-distribution results are in Tables~\ref{tab:std75} and~\ref{tab:complex}. Cross-wall: the test of \citet{zhou2024dino}, with start and goal on opposite sides of an unseen wall. $^{\ell}$: LeWM trained with single-layout batches. Bold marks the better of each pair.}
\label{tab:gen}
\setlength{\tabcolsep}{4pt}
\footnotesize
\footnotesize %
\scalebox{0.87}{%
\begin{tabular}{lcccccc}
\toprule
 & \multicolumn{2}{c}{\textbf{WallRandom}} & \multicolumn{2}{c}{\textbf{PushObj}} & \multicolumn{2}{c}{\textbf{TwoRoom}} \\
\cmidrule(lr){2-3}\cmidrule(lr){4-5}\cmidrule(lr){6-7}
\textbf{Planner}
& \textbf{\shortstack{unseen\\1 plan}}
& \textbf{\shortstack{unseen\\cross-wall}}
& \textbf{\shortstack{unseen\\1 plan}}
& \textbf{\shortstack{unseen\\3 plans}}
& \textbf{\shortstack{door moved\\1 plan}}
& \textbf{\shortstack{door moved\\3 plans}} \\
\midrule
LeWM$^{\ell}$ planner
& 67.5{\scriptsize$\pm$1.7}
& 15.1{\scriptsize$\pm$1.9}
& 53.6{\scriptsize$\pm$2.2}
& 18.8{\scriptsize$\pm$1.4}
& 65.3{\scriptsize$\pm$1.9}
& 9.9{\scriptsize$\pm$1.0} \\
\rowcolor{oursblue}
\quad \textbf{+ TEMPO (ours)}
& \textbf{76.9{\scriptsize$\pm$1.4}}
& \textbf{22.9{\scriptsize$\pm$2.5}}
& \textbf{54.8{\scriptsize$\pm$2.0}}
& \textbf{20.5{\scriptsize$\pm$1.4}}
& \textbf{74.5{\scriptsize$\pm$1.6}}
& \textbf{30.1{\scriptsize$\pm$0.8}} \\
\addlinespace[3pt]
PLDM planner
& 75.6{\scriptsize$\pm$2.0}
& 6.0{\scriptsize$\pm$1.0}
& 44.1{\scriptsize$\pm$1.6}
& 11.1{\scriptsize$\pm$0.8}
& 73.9{\scriptsize$\pm$1.7}
& 18.9{\scriptsize$\pm$1.0} \\
\rowcolor{oursblue}
\quad \textbf{+ TEMPO (ours)}
& \textbf{77.9{\scriptsize$\pm$2.0}}
& \textbf{6.5{\scriptsize$\pm$1.1}}
& \textbf{48.8{\scriptsize$\pm$1.7}}
& \textbf{15.3{\scriptsize$\pm$1.0}}
& \textbf{76.3{\scriptsize$\pm$1.5}}
& \textbf{30.4{\scriptsize$\pm$1.0}} \\
\bottomrule
\end{tabular}}
\end{table}

\subsection{Ablations}
\label{sec:ablation}
Appendix Table~\ref{tab:ablation} separates the two terms of the cost. $F_\phi$ alone remains competitive with the blend on TwoRoom and
Reacher but falls behind on Push-T for both planners and on Cube and Scene for LeWM, because the latent term carries the fine positioning that manipulation needs and $F_\phi$ the
ordering of far states, so combining the two terms works best. Finally, Appendix~\ref{sec:geometry} shows why the flat planner fails at range. Three plans out, the latent distance ranks where a plan ends but not
whether it succeeds, while \textit{TEMPO}'s cost ranks the same plans closer to their eventual success (Tables~\ref{tab:actionrank} and~\ref{tab:rank-tempo}).

\section{Conclusion}
\label{sec:conclusion}

We introduce \textit{TEMPO}, a simple yet effective temporal-distance planning objective that corrects the planning cost of frozen world models with one latent vector per state without retraining them, task-specific tuning or reward. A distance that counts environment steps, learned from the time steps of recorded trajectories and blended with the latent distance, turns the long-horizon collapse of flat planners into near-complete
success, taking LeWM from $36$ to $99\%$ on TwoRoom and from $90$ to $98\%$ on Reacher three plans from the goal and from $17$ to $75\%$
on the harder cross-room task. The same cost improves both the LeWM and the PLDM planner at every goal distance, including one plan away, for at most $0.3\%$ of a plan's arithmetic, and generalises without retraining to layout shifts (unseen wall layouts and door positions), shape shifts (unseen
block shapes), visual shifts (blur and colour jitter) and dynamics shifts (a larger agent or a heavier arm).

\paragraph{Future work.}
\label{sec:limitations}
Like the world models it builds on, \textit{TEMPO} benefits from a diverse offline dataset and follows the general trade-off that broader
coverage improves generalisation. It adds no requirement of its own, as $F_\phi$ is trained on the world model's own training set, needs no
reward signals or labels beyond the time steps of its trajectories, and works on random-walk data (Wall, WallRandom) as well as on demonstrations. Because
it keeps learning from the planner's own episodes, continuing to update it with new experience could enable progressive improvement without extensive interaction up front. Already, an $F_\phi$ trained only on training configurations substantially improves the planner on unseen wall
layouts and remains competitive on unseen block shapes. Another direction is to extend \textit{TEMPO} to world models that plan over the full spatial grid of patch tokens \citep{zhou2024dino}. Like the world models it corrects, \textit{TEMPO} is trained per environment with one shared recipe, and learning a single
temporal distance across environments on top of a multi-task world model is a further extension. Training $F_\phi$ toward the fewest steps between two states, rather than the steps a recorded trajectory took, for instance with an asymmetric quasimetric objective \citep{wang2023qrl}, is another natural extension. Finally, all our experiments are in simulation, and since \textit{TEMPO} leaves the world model untouched and learns only from the time steps of recorded trajectories, validating it on a real robot is a natural next step.

\bibliography{refs}
\bibliographystyle{plainnat}

\appendix
\section{Method details}
\label{app:method}

\paragraph{Learned temporal distance $F_\phi$.} MLP on the 192-d state latent with two hidden layers of 256 units (SiLU) to 64-d ($0.13$M parameters); loss of Eq.~\eqref{eq:fmap} on same-episode pairs
with $k \sim U[1, 75]$ ($\kappa = 25$) and a hinge of margin $3 = 75/\kappa$ on pairs more than 75 steps apart; Adam $10^{-3}$, batch 256, gradient clipping 10; 3,000
offline steps on the 20,000 expert windows, then one step per trainer update in every online round. Cost of Eq.~\eqref{eq:cost}: $\bar d_{25}$ is computed once per world model as the median norm of its 25-step latent displacement over its own encoded training data, so no scale is set by hand. Because $F_\phi$ keeps learning online
while the planner uses it, the embedding drifts between rounds; the planner re-solves from the current $F_\phi$ at every decision, so the drift is absorbed.

\paragraph{World model training.} Every world model follows the 10-epoch protocol of \citet{maes2026leworldmodel} on
stable-worldmodel \citep{maes2026stable}. A frame-skip of 5 groups the actions between two stored frames into one
action block. Training uses sub-trajectories of 4 frames, i.e.\ 3 context frames and 1 prediction, with 4 action blocks
of 5 actions, and the planner's horizon of 5 blocks is 25 environment steps. Frames are $224 \times 224$. LeWM's encoder
is a HuggingFace Vision Transformer Tiny with patch size 14, trained from scratch (no pretrained weights), emitting a
192-d class token; its predictor is a causal transformer over the frame history (depth 6, 16 heads, MLP width 2048, dropout 0.1)
that consumes the 192-d latents and the action blocks and emits the next latent. Batch size 128, AdamW at
$5\times10^{-5}$, three seeds (42, 43, 44) per environment, frozen after training. The Wall environment itself is the released planning environment of \citet{zhou2024dino}, in which one unit of action moves the dot by two arena units, whereas the released Wall dataset was generated at one unit per unit of action; every Wall evaluation here, as in the original, therefore executes the planner's actions at twice the displacement the world model was trained on. All arms share the environment, so the pairs in Table~\ref{tab:complex} are unaffected, but the absolute Wall numbers should be read with this in mind. PLDM trains its own encoder and predictor. Neither LeWM nor
PLDM receives any input beyond pixels and actions.

\paragraph{Planner.} stable-worldmodel CEM over LeWM's predictor: 300 samples, 30 iterations, top 30, horizon 5 action
blocks of 5 environment steps (actions StandardScaler-normalised in the planner's block layout), 3 context frames, warm
start from the previous plan, cost of Eq.~\eqref{eq:cost} on the last predicted latent (ours) or the latent distance alone (LeWM). Two settings differ from \citet{maes2026leworldmodel}, identically for every arm: the CEM runs 30 iterations in every environment (the
published setup uses 10 outside Push-T), and the predictor sees 3 context frames everywhere (1 on TwoRoom in the published setup).
The flat planner still reproduces the published LeWM numbers on LeWM's own protocol (Appendix~\ref{app:tables}).

\paragraph{Data and schedule.}
Every dataset frame is encoded once. Offline: 20,000 expert windows $[t_0, t_0 + \text{offset}]$ with $G = z_{t_0+\text{offset}}$,
3,000 gradient steps (Adam $10^{-3}$, batch 256). Online: 8 rounds of 50 training windows in 50 parallel environments; after each
decision batch $F_\phi$ takes 5 gradient steps on the stored same-episode pairs of every episode in the buffer (expert and the planner's
own), 50 at each episode end. \emph{Collection policy.} The online rounds of the reported cells are collected by an exploratory earlier planner variant that aims the same CEM at intermediate latents sampled around the current state; collecting them with the planner of Eq.~\eqref{eq:cost} itself gives the same success at goal 75 on TwoRoom, Push-T, Reacher, Cube and Scene, the largest difference over the five being $3.2$ points. Evaluation: $F_\phi$ frozen, 5 fixed groups of 50 windows per
world-model seed, three seeds, planner seed fixed, budget $2\times$ offset, success = the environment's criterion within budget. The
LeWM and ours arms of one cell are evaluated in one job on identical windows.

\section{Planning solver}
\label{app:cem}

\paragraph{Planning solver.} For planning we use the cross-entropy method (CEM) \citep{rubinstein2004cross}, the solver that performed
best in prior work on these world models \citep{maes2026leworldmodel,zhou2024dino}. At each planning step CEM samples 300 candidate action
sequences and optimises them for 30 iterations in every environment; at each iteration the top 30 sequences are retained to update the
sampling distribution, and the initial sampling variance is set to 1 in the normalised action space. The planning horizon is 5 steps,
which corresponds to 25 environment timesteps due to a frame skip of 5. We employ a receding-horizon model predictive control scheme
with a horizon of 5, meaning that the entire optimised action sequence is executed before replanning, and the next solve is
warm-started from the previous mean. This configuration follows the LeWM setup \citep{maes2026leworldmodel}, which uses 30 iterations on
Push-T and 10 elsewhere; we use 30 everywhere. The only difference between the arms of a pair is the cost $J$ in line 6 of
Algorithm~\ref{alg:cem}: the squared latent distance to the encoded goal for the LeWM and PLDM planners, and
Eq.~\eqref{eq:cost} for \textit{TEMPO}. LeWM and PLDM solve 50 environments per batch.

\begin{algorithm}[h]
\caption{Cross-entropy method over action sequences (one planning step)}
\label{alg:cem}
\begin{algorithmic}[1]
\Require frozen world model $\mathrm{WM}$, current latent $z_t$, goal $G$, horizon $H$, samples $N$, elites $K$, iterations $T$, cost $J$
\State initialise $\mu_0$ (previous plan, or $0$), $\sigma_0 = 1$
\For{$\tau = 1$ to $T$}
  \State sample $N$ action sequences $a^{(i)}_{1:H} \sim \mathcal{N}(\mu_{\tau-1}, \sigma_{\tau-1}^2)$
  \For{$i = 1$ to $N$}
    \State roll out $a^{(i)}_{1:H}$ through $\mathrm{WM}$ from $z_t$ to obtain $\hat z^{(i)}_{t+1:t+H}$; \quad $J^{(i)} \gets J(\hat z^{(i)}_{t+1:t+H}, G)$
  \EndFor
  \State $\mathcal{E} \gets$ the $K$ sequences with the lowest $J^{(i)}$
  \State $\mu_\tau \gets \frac{1}{K}\sum_{i\in\mathcal{E}} a^{(i)}_{1:H}$, \quad $\sigma^2_\tau \gets \mathrm{Var}_{i\in\mathcal{E}}\, a^{(i)}_{1:H}$
\EndFor
\State \Return $\mu_T$, executed for $H$ blocks before the next planning step
\end{algorithmic}
\end{algorithm}

\paragraph{Compute.} Both arms roll the same $300 \times 30$ candidates through the $10.8$M-parameter predictor for 5 steps each ($4.5 \times 10^{4}$ predictor steps per plan);
\textit{TEMPO} adds one evaluation of the $0.13$M-parameter $F_\phi$ per candidate, at most $0.3\%$ of the plan's arithmetic, and no second model or search.

\paragraph{Hyperparameters.} These settings are constant across environments and planners: every planner (LeWM and PLDM,
each with and without \textit{TEMPO}) uses 300 CEM samples, 30 iterations, the top 30 as elites and a 25-step plan in
every environment, and \textit{TEMPO} blends its temporal distance with weight $w = 0.5$ everywhere. Goals lie 25, 50 and 75 steps
ahead (one to three plans); Wall, whose episodes end at 50 steps, uses 25 and 40. Each cell is evaluated on five fixed groups of 50
windows per world-model seed, three seeds.

\section{Environments and datasets}
\label{app:envs}

\begin{figure}[h]
\centering
\includegraphics[width=\textwidth]{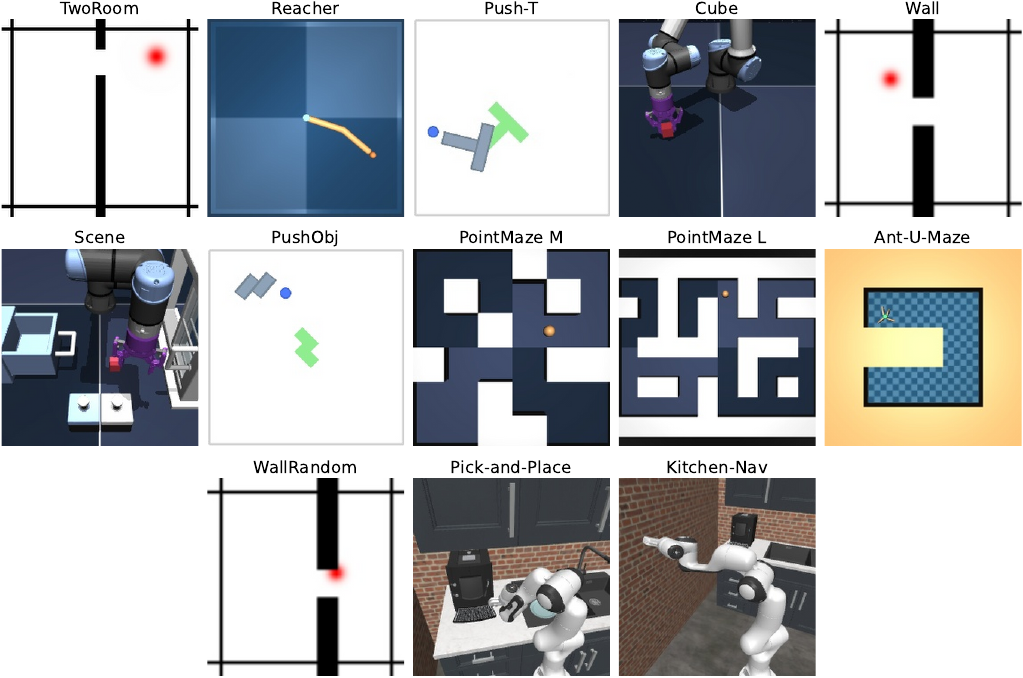}
\caption{\textbf{The eleven environments and two RoboCasa kitchen tasks}, left to right, top to bottom: TwoRoom, Reacher, Push-T, Cube, Wall, Scene, PushObj,
PointMaze medium and large, Ant-U-Maze, WallRandom, and the RoboCasa kitchen tasks Pick-and-Place and
Kitchen-Nav. Each panel is one $224\times224$ training frame, the only observation the world models and the planner receive.}
\label{fig:envs}
\end{figure}

Figure~\ref{fig:envs} shows one training frame of each environment. Frames are $224 \times 224$ RGB, actions are grouped into blocks of 5 by a frame-skip of 5, and every world model is trained for
10 epochs on the sets below (Appendix~\ref{app:method}); Table~\ref{tab:datasets} gives their sizes. Success is each environment's
own criterion.

\begin{table}[h]
\centering\small
\caption{\textbf{Training sets.} Episodes the world models were trained on, mean episode length in environment steps, and action
dimension. Where a 90:10 episode split defines the protocol (Section~\ref{sec:setup}) the count is the 90\% training split.}
\label{tab:datasets}
\setlength{\tabcolsep}{5pt}
\begin{tabular}{lrrr}
\toprule
\textbf{Environment} & \textbf{Episodes (trained on)} & \textbf{Mean length} & \textbf{Action dim.} \\
\midrule
TwoRoom & 10,000 & 92 & 2 \\
Push-T & 18,685 & 125 & 2 \\
Reacher & 10,000 & 200 & 2 \\
Cube & 10,000 & 200 & 5 \\
Scene & 10,000 & 200 & 5 \\
PushObj & 20,000 & 101 & 2 \\
Wall & 1,920 & 50 & 2 \\
PointMaze medium & 2,700 & 200 & 2 \\
PointMaze large & 2,700 & 200 & 2 \\
Ant-U-Maze & 6,467 & 100 & 8 \\
RoboCasa Pick-and-Place & 1,935 & 300 & 6 \\
RoboCasa Kitchen-Nav & 1,789 & 300 & 3 \\
\bottomrule
\end{tabular}
\end{table}

\begin{enumerate}[label=(\alph*),leftmargin=*]
\item \textbf{TwoRoom} is the continuous 2D navigation task of \citet{sobal2026learning}: two rooms separated by a wall with a single door.
The agent, a coloured dot, must travel from a random start in one room to a randomly sampled goal in the other, which requires passing
through the door. We use the 10,000 episodes of the PLDM/LeWM release, average length 92 steps, generated by a noisy heuristic that
drives the agent to the door and then to the goal. Each world model is trained on this set for 10 epochs.
\item \textbf{Push-T} is the continuous 2D manipulation task in which an agent (a blue dot) must push a T-shaped block to a target
pose through pushing contacts only. We follow the setup and expert dataset of \citet{zhou2024dino} through the platform's training
split, 18,685 episodes of average length 125 steps, and, as LeWM does, train each world model for 10 epochs.
\item \textbf{Reacher} is the two-joint arm of the DeepMind Control Suite \citep{tassa2018deepmind}. Following \citet{zhou2024dino}, success requires the
arm to match the joint configuration that reaches the target. The platform's 10,000 episodes of 200 steps are collected with a Soft
Actor-Critic policy; each world model is trained on them for 10 epochs.
\item \textbf{Cube} and \textbf{Scene} are the OGBench single-object manipulation tasks \citep{park2025ogbench}: a robot arm must move a
cube to a target pose (Cube), or operate a drawer, a window, two buttons and a cube into a target configuration (Scene). For each we
collect 10,000 episodes of 200 steps with the benchmark's scripted expert and train each world model for 10 epochs.
\item \textbf{PushObj} is Push-T with the block shape varied, from the generalisation suite of \citet{zhou2024dino}. We
collect 20,000 episodes of 100 steps over the four training shapes (T, o, square, small tee) with the platform's Push-T collection
policy, and hold out the original test shapes for Table~\ref{tab:gen}; success requires both the agent and the block at their targets.
\item \textbf{Wall} is the point agent of \citet{zhou2024dino} that must cross a wall through its door. We use their 1,920 random-walk trajectories of
50 steps; because episodes end at 50 steps its horizons are 25 and 40 steps rather than 25, 50 and 75. \textbf{WallRandom}
(Table~\ref{tab:gen}) is the same task with the wall and door positions randomised, 10,240 random-walk trajectories of 50 steps for
training and held-out layouts for evaluation, as in the original.
\item \textbf{PointMaze medium} and \textbf{large} are the OGBench mazes with the benchmark's expert trajectories, 3,000 episodes of 200
steps each, split 90:10 by episode; the world models train on the 2,700 training episodes for 10 epochs.
\item \textbf{Ant-U-Maze} is D4RL's antmaze-umaze-diverse \citep{fu2020d4rl}, an 8-d quadruped in a U-shaped maze. The recorded action
sequences are re-simulated in our simulator in 100-step chunks so that every stored state is reproducible at evaluation
(Appendix~\ref{app:antu}), giving 6,467 training and 698 held-out chunk-episodes of 100 steps.
\item \textbf{RoboCasa Pick-and-Place} and \textbf{Kitchen-Nav} are two tasks in a RoboCasa kitchen \citep{nasiriany2024robocasa} with a
mobile manipulator: moving an object between the counter and the sink, and driving the base to a target pose. Pick-and-Place succeeds
when the object lies within $0.05$\,m of its goal position; Kitchen-Nav when the base is within $0.20$\,m of the goal (the task's own
threshold) with a heading cosine of at least $0.95$ (the task uses $0.98$). Both are collected by our scripted experts. The
Pick-and-Place expert repeatedly picks the object and places it, alternating between the sink and the counter, driving the mobile base
into reach before each approach and carry; it retries a failed grasp up to three times and then falls back to random arm motion for
the rest of the episode. The Kitchen-Nav expert is a waypoint follower over a floor grid rasterised from the fixtures' footprints. The
scene of each episode is stored so that evaluation rebuilds it exactly. Pick-and-Place has
1,935 episodes of 300 steps (2,000 collected, 65 dropped for render faults) and Kitchen-Nav 1,988 episodes of 300 steps split 90:10;
each world model is trained for 10 epochs. Windows already solved at their start row are excluded for every arm before the seeded
draw: on Pick-and-Place those whose object lies within $0.05$\,m of its goal ($332{,}570$, $227{,}350$ and $159{,}342$ of $582{,}435$
rows at one, two and three plans), on Kitchen-Nav those whose base lies within $0.20$\,m ($17{,}301$, $12{,}038$ and $8{,}360$ of
$59{,}899$).
\end{enumerate}

\paragraph{Ant-U-Maze.}
\label{app:antu}
Ant-U is the Ant-U-Maze of \citet{sobal2026learning} (D4RL antmaze-umaze-diverse), planned from pixels. D4RL's recorded states come from a different simulator, so
we re-simulate its recorded actions in our MuJoCo environment with D4RL's ant model, in 100-step chunks anchored at the recorded states,
dropping chunks that start with the ant on its back or stuck; the split is by source episode (6{,}467 training and 698 held-out chunks).
An evaluation window needs the goal at least 1~m away, the ant upright at the start and the goal, and a goal the recorded actions reach.
Success is the maze goal within $0.5$~m, as in \citet{sobal2026learning}, whose Ant-U model is state-based, so this column is an internal
comparison, not a reproduction.

\section{Baseline: PLDM as trained here}
\label{app:baselines}

The PLDM baseline is trained on the stable-worldmodel platform \citep{maes2026stable} with the configurations released
there, for the same 10 epochs, on the same recorded trajectories, frame-skip and $224\times224$ frames as the LeWM boards
(Appendix~\ref{app:method}), one board per seed (42, 43, 44). We did not tune it per environment; every coefficient below is the
platform's default, and the same values were used on every environment. Its planner is the published one: sampling-based MPC over
the frozen model with the squared latent distance to the encoded goal as the cost.

\paragraph{PLDM.}
PLDM \citep{sobal2026learning} learns encoder and predictor end to end and prevents collapse with a VICReg-style regulariser
\citep{bardes2021vicreg} extended along time. The objective is a weighted sum of a prediction term and six regularisers,
\begin{equation}
  \mathcal{L}_{\text{PLDM}} = \mathcal{L}_{\text{pred}} + \alpha\,\mathcal{L}_{\text{var}} + \beta\,\mathcal{L}_{\text{cov}}
     + \gamma\,\mathcal{L}_{\text{time-sim}} + \zeta\,\mathcal{L}_{\text{time-var}} + \nu\,\mathcal{L}_{\text{time-cov}} + \mu\,\mathcal{L}_{\text{IDM}}, \label{eq:pldm-loss}
\end{equation}
with each term as defined by \citet{sobal2026learning}.
The variance and covariance terms hold the batch of latents at each step spread out and decorrelated; their temporal counterparts do the
same along each trajectory; the time-similarity term keeps consecutive latents close; the inverse-dynamics term asks the pair
$(z_t, z_{t+1})$ to determine the action. \citet{maes2026leworldmodel} report that the original coefficients were tuned per environment and
dataset and re-tuned them by a grid search over 256 configurations on Push-T. What we did instead: we kept the platform's released
coefficients for every environment, with no search: $\alpha = 18$, $\beta = 12$, $\gamma = 0.2$, $\zeta = 0.7$ and $\nu = \mu = 0$
(the temporal covariance and inverse-dynamics terms are present in the code but switched off), a projector loss of weight $10^{-3}$,
and no SIGReg term. The encoder is a ViT-Ti/14 trained from scratch with a 192-d latent through a two-layer projector (hidden width 2048,
batch norm); the predictor is a transformer of depth 6, 16 heads, MLP width 2048, dropout 0.1 over a three-frame history predicting one
frame; optimisation is AdamW at learning rate $5\times10^{-5}$ and weight decay $10^{-3}$, batch size 128, bf16, gradient clipping at 1.0,
10 epochs. Its planning cost is the published squared latent distance, run through the same CEM as LeWM's
(Appendix~\ref{app:portability}). Because no coefficient was tuned on our environments, the PLDM rows of Table~\ref{tab:std75} are a
conservative baseline: a per-environment search of the kind \citet{maes2026leworldmodel} ran could raise them, and it would raise the PLDM $+F_\phi$
row with them, since $F_\phi$ reads whatever latent the board provides.

\section{The method on other world models: what $F_\phi$ asks of them}
\label{app:portability}
The method is the cost, not the planner it was first measured on: every ``$+\,F_\phi$'' row keeps the planner above it and
changes only what that planner minimises. What follows is therefore not the portability of one system but the same method applied
to a second frozen world model. The learned temporal distance touches a world model through two calls and nothing else. \emph{An encoder} that maps one frame to
one latent vector $z$; for LeWM this is the 192-d projected class token of its ViT, with no proprioception. \emph{A rollout} that
takes the context latents and a candidate action sequence and returns the predicted latents along the plan; the CEM of
Eq.~\eqref{eq:cost} scores its last entry. $F_\phi$ is the same MLP with its input width set to the latent dimension, its loss
(Eq.~\eqref{eq:fmap}) reads only the time steps of the recorded trajectories, and the blend of Eq.~\eqref{eq:cost} needs the model's own distance as its first
term. Any frozen latent world model that exposes those two calls, one latent vector per frame and a latent rollout, can carry it.

\paragraph{PLDM.} PLDM's model exposes the same two calls with the same shapes: a ViT class token through a projector to a 192-d
latent, and an autoregressive predictor over a three-frame history, trained on the platform with the same data and schedule as the LeWM boards. The two differ
in the training objective, an inverse-dynamics term with a variance--covariance regulariser for PLDM against LeWM's SIGReg, not in the
interface. Running Eq.~\eqref{eq:cost} on a PLDM board is therefore the same harness with the checkpoint swapped; the flat planner
on the same board, PLDM's published latent-distance cost run through the same CEM, is the matched control. The results on PLDM are the PLDM rows of Tables~\ref{tab:std75} and~\ref{tab:complex}.

\paragraph{Why the baseline rows use the published costs.} PLDM enters Table~\ref{tab:std75} as
the planner its authors published, run through the same CEM on the same windows; that is what a baseline row should mean. The
question this section answers is a different one, whether the temporal distance is a property of the cost rather than of LeWM's
latent, and it needs the matched pairs above to be settled.

\section{Ablations and per-offset results}
\label{app:tables}

\paragraph{Reproducing the published LeWM planner.} Run on LeWM's own protocol, our flat planner matches the published
numbers of \citet{maes2026leworldmodel} on TwoRoom ($87.7$ against $88.8$), Push-T ($93.6$ against $90.4$) and Cube ($74.9$
against $72.4$), and exceeds them on Reacher ($94.7$ against $81.2$). The Reacher gap is the CEM budget rather than the
cost: we run 30 CEM iterations in every environment, where the published setup uses 10 outside Push-T.

Table~\ref{tab:ablation} separates the two terms of the cost (Section~\ref{sec:ablation}). With the world model and planner of each
row fixed, it compares the blend of Eq.~\eqref{eq:cost} with $F_\phi$ alone on the five environments of Table~\ref{tab:std75}, one to three
plans from the goal, for LeWM and PLDM; the blend has the higher average in every block.

\begin{table}[t]
\centering\small
\caption{\textbf{Ablation of the planning cost on each frozen world model.} The world model and planner are fixed and only the cost changes: latent $+\,F_\phi$ is the cost of Eq.~\eqref{eq:cost} used in every other table, and $F_\phi$ alone drops the latent term ($w = 1$). Bold marks the better of each pair. Avg.: mean over the five environments. }
\label{tab:ablation}
\setlength{\tabcolsep}{4pt}
\footnotesize %
\scalebox{0.75}{%
\begin{tabular}{lcccccc}
\toprule
\textbf{Ablation } & \textbf{TwoRoom} & \textbf{Push-T} & \textbf{Reacher} & \textbf{Cube} & \textbf{Scene} & \textbf{Avg.} \\
\midrule
\multicolumn{7}{l}{\textbf{LeWM}} \\
\rowcolor{gray!12}\multicolumn{7}{l}{\textit{One plan from the goal}} \\
$F_\phi$ alone ($w = 1$) & \textbf{98.3{\scriptsize$\pm$0.3}} & 89.1{\scriptsize$\pm$0.9} & 93.5{\scriptsize$\pm$0.6} & 75.9{\scriptsize$\pm$2.1} & 68.5{\scriptsize$\pm$1.5} & 85.0 \\
\rowcolor{oursblue}\textbf{Latent $+\,F_\phi$ (ours, $w = 0.5$)} & 97.2{\scriptsize$\pm$0.6} & \textbf{95.1{\scriptsize$\pm$0.6}} & \textbf{95.5{\scriptsize$\pm$0.7}} & \textbf{77.1{\scriptsize$\pm$1.7}} & \textbf{71.1{\scriptsize$\pm$1.1}} & \textbf{87.2} \\
\rowcolor{gray!12}\multicolumn{7}{l}{\textit{Two plans}} \\
$F_\phi$ alone ($w = 1$) & 97.2{\scriptsize$\pm$0.7} & 30.4{\scriptsize$\pm$1.9} & 96.5{\scriptsize$\pm$0.5} & 52.3{\scriptsize$\pm$2.1} & 52.5{\scriptsize$\pm$1.5} & 65.8 \\
\rowcolor{oursblue}\textbf{Latent $+\,F_\phi$ (ours, $w = 0.5$)} & \textbf{99.2{\scriptsize$\pm$0.3}} & \textbf{55.7{\scriptsize$\pm$1.2}} & \textbf{96.8{\scriptsize$\pm$0.5}} & \textbf{58.7{\scriptsize$\pm$2.0}} & \textbf{57.9{\scriptsize$\pm$1.3}} & \textbf{73.7} \\
\rowcolor{gray!12}\multicolumn{7}{l}{\textit{Three plans}} \\
$F_\phi$ alone ($w = 1$) & 99.1{\scriptsize$\pm$0.3} & 16.0{\scriptsize$\pm$1.2} & 97.9{\scriptsize$\pm$0.4} & 67.1{\scriptsize$\pm$1.9} & 47.9{\scriptsize$\pm$1.4} & 65.6 \\
\rowcolor{oursblue}\textbf{Latent $+\,F_\phi$ (ours, $w = 0.5$)} & \textbf{99.2{\scriptsize$\pm$0.3}} & \textbf{28.1{\scriptsize$\pm$2.0}} & \textbf{98.4{\scriptsize$\pm$0.5}} & \textbf{70.4{\scriptsize$\pm$1.2}} & \textbf{50.5{\scriptsize$\pm$1.3}} & \textbf{69.3} \\
\addlinespace[4pt]
\midrule
\multicolumn{7}{l}{\textbf{PLDM}} \\
\rowcolor{gray!12}\multicolumn{7}{l}{\textit{One plan from the goal}} \\
$F_\phi$ alone ($w = 1$) & 98.7{\scriptsize$\pm$0.4} & 82.4{\scriptsize$\pm$1.3} & 94.1{\scriptsize$\pm$0.8} & \textbf{71.1{\scriptsize$\pm$2.0}} & \textbf{61.6{\scriptsize$\pm$1.7}} & 81.6 \\
\rowcolor{oursblue}\textbf{Latent $+\,F_\phi$ (ours, $w = 0.5$)} & \textbf{98.9{\scriptsize$\pm$0.3}} & \textbf{90.1{\scriptsize$\pm$1.2}} & \textbf{94.5{\scriptsize$\pm$1.0}} & 68.7{\scriptsize$\pm$1.4} & 58.4{\scriptsize$\pm$1.3} & \textbf{82.1} \\
\rowcolor{gray!12}\multicolumn{7}{l}{\textit{Two plans}} \\
$F_\phi$ alone ($w = 1$) & 96.1{\scriptsize$\pm$1.0} & 27.2{\scriptsize$\pm$1.4} & 96.9{\scriptsize$\pm$0.6} & 43.5{\scriptsize$\pm$2.1} & \textbf{46.4{\scriptsize$\pm$1.4}} & 62.0 \\
\rowcolor{oursblue}\textbf{Latent $+\,F_\phi$ (ours, $w = 0.5$)} & \textbf{97.2{\scriptsize$\pm$0.5}} & \textbf{39.5{\scriptsize$\pm$1.8}} & \textbf{98.0{\scriptsize$\pm$0.4}} & \textbf{48.7{\scriptsize$\pm$1.8}} & 45.9{\scriptsize$\pm$1.2} & \textbf{65.8} \\
\rowcolor{gray!12}\multicolumn{7}{l}{\textit{Three plans}} \\
$F_\phi$ alone ($w = 1$) & 94.7{\scriptsize$\pm$1.3} & 11.5{\scriptsize$\pm$1.0} & \textbf{97.9{\scriptsize$\pm$0.2}} & 61.6{\scriptsize$\pm$2.2} & \textbf{43.5{\scriptsize$\pm$1.1}} & 61.8 \\
\rowcolor{oursblue}\textbf{Latent $+\,F_\phi$ (ours, $w = 0.5$)} & \textbf{95.7{\scriptsize$\pm$1.1}} & \textbf{17.1{\scriptsize$\pm$1.7}} & 97.7{\scriptsize$\pm$0.5} & \textbf{66.7{\scriptsize$\pm$2.0}} & 41.1{\scriptsize$\pm$1.5} & \textbf{63.7} \\
\bottomrule
\end{tabular}}
\end{table}

\paragraph{Two plans away.}
At offset 50 (Table~\ref{tab:std75}, middle block) the order is the same as at three plans, with smaller gaps on the environments the flat
planner still solves: TwoRoom from $58.5$ to $99.2$, Push-T from $49.7$ to $55.7$, Reacher from $95.7$ to $96.8$, Cube from $57.9$ to $58.7$
and Scene from $56.4$ to $57.9$. Against the published PLDM planner, \textit{TEMPO} improves every environment at two plans.

\section{Generalisation to unseen configurations and test-time shifts}
\label{app:shift}
Every evaluation here changes the environment rather than the world model, which stays frozen: unseen layouts and block shapes
(Table~\ref{tab:gen}), the cross-room task, pixel corruptions and factor shifts of the scene and its dynamics.

\paragraph{Generalisation suites (WallRandom, PushObj).}
The generalisation study of \citet{zhou2024dino} retrains the world model on randomised wall and door layouts (WallRandom) and on several
block shapes (PushObj) and tests on layouts and shapes never seen in training. We rebuild both suites on the platform with our recipe
(10 training epochs, the same planner settings as Table~\ref{tab:std75}) and report, per world model, the planner with its own cost against
the same planner $+\,F_\phi$, where $F_\phi$ is trained on the training layouts or shapes only. WallRandom: 10{,}240 random-walk episodes over
1{,}044 layouts; the 81 held-out layouts have wall and door positions outside every training range. PushObj: 20{,}000 episodes with the
shapes T, Z, square and small tee; held-out shapes L, I and +. Figure~\ref{fig:gen-setups} shows the training and test setups, and Table~\ref{tab:gen} in the main text
gives the training and held-out columns.

uses the standard 25-step windows drawn from held-out episodes on the unseen layouts, `cross-wall' is the original start-on-one-side,
goal-on-the-other test. LeWM's WallRandom world model ($^{\ell}$) is trained with single-layout batches: every batch of 128 clips is drawn
from one of the 1{,}044 training layouts, so that the batch-level SIGReg regulariser has to be satisfied by the agent's motion rather than by
layout identity. Everything
else, including the 10-epoch budget, is the shared recipe; the planner and $+\,F_\phi$ rows share the world model.

\paragraph{Cross-wall diagnostic.}
\label{app:crosswall}
On the cross-wall test every arm of every 10-epoch board trained with mixed-layout batches is near zero on held-out layouts, and also on
training layouts, so for those boards the test measures a limitation of the planners rather than layout generalisation (the single-layout-batch
LeWM board of Table~\ref{tab:gen} is the exception). Pinning a single training layout (the fixed Wall
layout) and the same 250 cross-wall trials for every board separates the two levers, training length and cost. The 10-epoch fixed-layout LeWM
board crosses in $19.6 \pm 2.0\%$ of trials with its own cost and in $20.0 \pm 7.3\%$ with $+\,F_\phi$, the 10-epoch PLDM board in
$13.2 \pm 3.0\%$ and $11.2 \pm 2.0\%$, the 10-epoch WallRandom LeWM board in $0.4\%$. Trained for 100 epochs (the budget of \citet{zhou2024dino}), the fixed-layout board planned with its own cost reaches
$0.8$, $14.0$, $13.2$, $19.6$ and $21.2\%$ at epochs 20, 40, 60, 80 and 100, and its mean final distance to the goal stays above 30 of the
37.5 arena units between start and goal: the planner rarely leaves its side of the wall. The same 100-epoch boards planned with $+\,F_\phi$
cross in $74.3 \pm 3.3\%$ of the identical trials against $19.5 \pm 1.3\%$ for their own cost (three seeds, $74.0$, $78.4$ and $70.4$ against
$21.2$, $18.0$ and $19.2$), with a mean final distance of 13 units. The latent-distance cost does not find the
door at any training length; the temporal metric does once the world model's rollouts are accurate enough for it to rank them. Table~\ref{tab:gen}
reports the 10-epoch boards of the shared recipe; the 100-epoch WallRandom boards are trained with the same recipe otherwise.

\paragraph{Cross-room TwoRoom.} A harder change of the task rather than of the pixels: the Two-Rooms evaluation of
\citet{sobal2026learning} places the start in one room and the goal in the other, so every trial must pass through the door (50 identical start and goal
pairs per group, five groups per seed, three seeds, the same for every arm), and the
planner must commit to a route that moves away from the goal before it approaches it. Success is the environment's own criterion,
within 16 px of the goal, and the world models, planner and $F_\phi$ are those of the TwoRoom column of Table~\ref{tab:std75},
with $F_\phi$ not retrained. The cross-room column of Table~\ref{tab:complex} reports it at budgets of 50, 100 and 150 steps; the
50-step budget is tight, since the median successful traversal takes 48 steps, which is why both arms score lowest there. The gain
is the largest in this paper: the latent distance alone cannot express a route that first increases it, and the temporal distance can.

\begin{figure}[t]
\centering
\includegraphics[width=\textwidth]{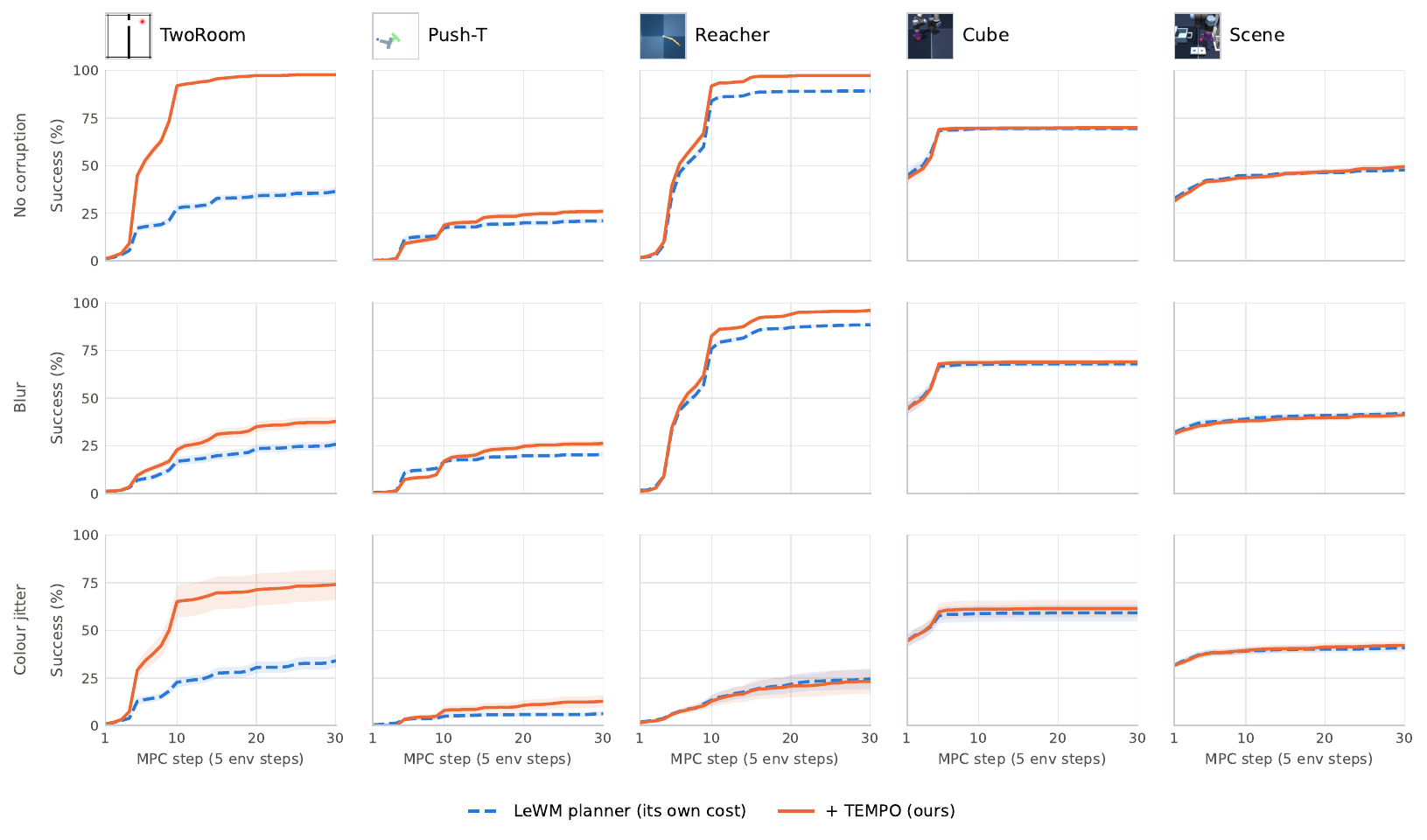}
\caption{\textbf{Test-time pixel corruptions: cumulative success versus MPC step, goal 75 steps ahead (30 steps = budget 150).}
Rows: condition; columns: environment. Dashed blue: the LeWM planner with its own cost; orange: the same planner $+$ \textit{TEMPO}
(ours), on identical windows.}
\label{fig:corruption-curves}
\end{figure}

\paragraph{Pixel corruptions.}
\label{app:corruption}
We keep the Table~\ref{tab:std75} world models unchanged and apply one of two generic
corruptions to every rendered frame at test time (blur, colour jitter; parameters as in \citealp{wang2026adajepa}); the
world model is frozen, $F_\phi$ is trained in the training environment and kept fixed during evaluation, and the
reference row runs the same setup with no corruption. Table~\ref{tab:corruption} gives the per-environment numbers (goal 75 steps ahead, 5 groups $\times$ 3 seeds).
\textit{TEMPO} lifts the average under every condition, from $52.8$ to $68.3$ with no corruption, $49.0$ to $54.1$ under blur and $33.0$ to
$42.8$ under colour jitter, and it doubles the planner on TwoRoom under colour jitter. Figure~\ref{fig:corruption-curves} shows the same runs as success over the episode. Table~\ref{tab:corruption-complex} applies the same
corruptions to the environments of Table~\ref{tab:complex}; \textit{TEMPO} raises the average under every condition.

\begin{table}[t]
\centering\small
\caption{\textbf{Same out-of-distribution protocol on every environment (goal 75 steps ahead).} The pixel corruptions are generic wrappers on the rendered frame, applied with identical parameters in every environment (values as in \citealp{wang2026adajepa}); the reference row runs the same protocol with no corruption, so the first frame is re-rendered live in every row. World model frozen, $F_\phi$ trained online in the corrupted environment, blend cost of Eq.~\eqref{eq:cost}. Bold: the better of each pair.}
\label{tab:corruption}
\setlength{\tabcolsep}{4pt}
\footnotesize %
\scalebox{0.75}{%
\begin{tabular}{lcccccc}
\toprule
\textbf{Condition} & \textbf{TwoRoom} & \textbf{Push-T} & \textbf{Reacher} & \textbf{Cube} & \textbf{Scene} & \textbf{Avg.} \\
\midrule
in-distribution — LeWM planner & 36.5{\scriptsize$\pm$2.1} & 21.1{\scriptsize$\pm$1.3} & 89.2{\scriptsize$\pm$0.5} & 69.5{\scriptsize$\pm$1.3} & 47.7{\scriptsize$\pm$1.0} & 52.8 \\
\rowcolor{oursblue}in-distribution — \textbf{+ TEMPO (ours)} & \textbf{98.4{\scriptsize$\pm$0.3}} & \textbf{26.1{\scriptsize$\pm$1.4}} & \textbf{97.7{\scriptsize$\pm$0.4}} & \textbf{70.0{\scriptsize$\pm$1.0}} & \textbf{49.5{\scriptsize$\pm$1.7}} & \textbf{68.3} \\
\midrule
blur — LeWM planner & 25.9{\scriptsize$\pm$2.2} & 20.7{\scriptsize$\pm$1.1} & 88.5{\scriptsize$\pm$0.5} & 68.0{\scriptsize$\pm$1.6} & \textbf{42.0{\scriptsize$\pm$1.4}} & 49.0 \\
\rowcolor{oursblue}blur — \textbf{+ TEMPO (ours)} & \textbf{38.1{\scriptsize$\pm$2.4}} & \textbf{26.5{\scriptsize$\pm$1.7}} & \textbf{96.0{\scriptsize$\pm$0.6}} & \textbf{68.9{\scriptsize$\pm$1.5}} & 40.5{\scriptsize$\pm$1.5} & \textbf{54.0} \\
\midrule
colour jitter — LeWM planner & 34.1{\scriptsize$\pm$3.3} & 6.3{\scriptsize$\pm$1.3} & \textbf{24.5{\scriptsize$\pm$5.2}} & 59.2{\scriptsize$\pm$4.3} & 41.1{\scriptsize$\pm$2.1} & 33.0 \\
\rowcolor{oursblue}colour jitter — \textbf{+ TEMPO (ours)} & \textbf{74.3{\scriptsize$\pm$8.0}} & \textbf{12.8{\scriptsize$\pm$3.2}} & 23.2{\scriptsize$\pm$6.5} & \textbf{61.5{\scriptsize$\pm$4.5}} & \textbf{42.1{\scriptsize$\pm$2.2}} & \textbf{42.8} \\
\bottomrule
\end{tabular}}
\end{table}

\begin{table}[t]
\centering\small
\caption{\textbf{The same pixel-corruption protocol on the more complex environments of Table~\ref{tab:complex} (goal 75 steps ahead).} The pixel corruptions are generic wrappers on the rendered frame, applied with identical parameters in every environment (values as in \citealp{wang2026adajepa}); the in-distribution rows repeat the corresponding cells of Table~\ref{tab:complex}. $^\dagger$Wall episodes end at 50 steps, so its goal is 40 steps ahead; cross-room uses its 150-step budget. World model frozen, $F_\phi$ trained online in the corrupted environment, blend cost of Eq.~\eqref{eq:cost}. Bold: the better of each pair.}
\label{tab:corruption-complex}
\setlength{\tabcolsep}{4pt}
\footnotesize %
\scalebox{0.75}{%
\begin{tabular}{lcccccc}
\toprule
\textbf{Condition} & \textbf{Wall$^\dagger$} & \textbf{\shortstack{TwoRoom\\cross-room}} & \textbf{PointMaze M} & \textbf{PointMaze L} & \textbf{Ant-U} & \textbf{Avg.} \\
\midrule
in-distribution — LeWM planner & 57.7{\scriptsize$\pm$2.4} & 17.1{\scriptsize$\pm$1.1} & 10.8{\scriptsize$\pm$0.8} & 0.3{\scriptsize$\pm$0.2} & 28.1{\scriptsize$\pm$1.6} & 22.8 \\
\rowcolor{oursblue}in-distribution — \textbf{+ TEMPO (ours)} & \textbf{65.9{\scriptsize$\pm$2.2}} & \textbf{74.5{\scriptsize$\pm$1.7}} & \textbf{31.7{\scriptsize$\pm$1.0}} & \textbf{6.5{\scriptsize$\pm$1.0}} & \textbf{34.0{\scriptsize$\pm$2.1}} & \textbf{42.5} \\
\midrule
blur — LeWM planner & 52.1{\scriptsize$\pm$2.8} & 8.9{\scriptsize$\pm$1.4} & 10.4{\scriptsize$\pm$0.9} & 0.1{\scriptsize$\pm$0.1} & \textbf{12.5{\scriptsize$\pm$1.4}} & 16.8 \\
\rowcolor{oursblue}blur — \textbf{+ TEMPO (ours)} & \textbf{60.9{\scriptsize$\pm$2.4}} & \textbf{23.3{\scriptsize$\pm$2.5}} & \textbf{17.9{\scriptsize$\pm$1.7}} & \textbf{4.1{\scriptsize$\pm$1.0}} & 12.4{\scriptsize$\pm$1.5} & \textbf{23.7} \\
\midrule
colour jitter — LeWM planner & 56.4{\scriptsize$\pm$2.6} & 16.1{\scriptsize$\pm$1.5} & 9.6{\scriptsize$\pm$1.0} & 1.6{\scriptsize$\pm$0.6} & 17.7{\scriptsize$\pm$1.7} & 20.3 \\
\rowcolor{oursblue}colour jitter — \textbf{+ TEMPO (ours)} & \textbf{68.0{\scriptsize$\pm$2.4}} & \textbf{59.7{\scriptsize$\pm$6.9}} & \textbf{10.4{\scriptsize$\pm$1.3}} & \textbf{1.7{\scriptsize$\pm$0.7}} & \textbf{19.7{\scriptsize$\pm$1.3}} & \textbf{31.9} \\
\bottomrule
\end{tabular}}
\end{table}

\paragraph{Factor shifts.} We follow an established factor-shift protocol \citep{wang2026adajepa}: the first frame is rendered live under the shift, the goal
frame stays the clean dataset frame, $F_\phi$ is trained in the training environment and kept fixed during evaluation, and the world model stays frozen; goal 75
steps ahead. Table~\ref{tab:env-shift} reports geometry, dynamics and camera shifts per environment. Where the shift leaves the scene legible to the frozen encoder, \textit{TEMPO} keeps its gain: on
Push-T a larger agent goes from $18.3$ to $26.3$, on TwoRoom a larger agent goes from $36.4$ to $94.9$ and a moved door from $9.9$ to $30.3$, and on Reacher a heavier arm goes from $89.7$ to $97.2$ (Table~\ref{tab:env-shift}). On Cube, where \textit{TEMPO} remains competitive with the planner without a shift, it remains so under the shifts as well.

\begin{table}[h]
\centering\small
\caption{\textbf{Factor shifts on Push-T, TwoRoom, Reacher, Cube and Scene, goal 75 steps ahead.} Protocol of \citet{wang2026adajepa}: the first frame is rendered live under the shift, the goal frame stays the clean dataset frame, the world model is frozen and $F_\phi$ is trained online in the shifted environment. Shifts change the geometry, the dynamics or the camera. Success (\%), LeWM planner and the same planner with \textit{TEMPO}; mean over three seeds; bold: the better of the pair.}
\label{tab:env-shift}
\footnotesize %
\begin{tabular}{llcc}
\toprule
\textbf{Environment} & \textbf{Shift} & \textbf{Planner} & \textbf{+ \textit{TEMPO}} \\
\midrule
Push-T & none (in distribution) & 21.1 & \textbf{26.1} \\
 & larger agent & 18.3 & \textbf{26.3} \\
\addlinespace[3pt]
TwoRoom & none (in distribution) & 36.5 & \textbf{98.4} \\
 & door moved & 9.9 & \textbf{30.3} \\
 & larger agent & 36.4 & \textbf{94.9} \\
\addlinespace[3pt]
Reacher & none (in distribution) & 89.2 & \textbf{97.7} \\
 & heavier arm & 89.7 & \textbf{97.2} \\
\addlinespace[3pt]
Cube & none (in distribution) & 69.5 & \textbf{70.0} \\
 & larger cube & 66.4 & \textbf{67.6} \\
 & camera tilted & 60.8 & \textbf{61.1} \\
\addlinespace[3pt]
Scene & none (in distribution) & 47.7 & \textbf{49.5} \\
 & larger cube & 46.5 & \textbf{48.4} \\
 & camera tilted & 35.9 & \textbf{36.9} \\
\bottomrule
\end{tabular}
\end{table}

\clearpage   %
\section{Additional results on RoboCasa kitchen tasks}
\label{app:robocasa}
\label{sec:robocasa}
We additionally evaluate on two RoboCasa kitchen tasks with a mobile manipulator (Table~\ref{tab:robocasa}), Pick-and-Place and Kitchen-Nav, with
world models trained on our own scripted-expert collections. These tasks are harder than the core suite for every planner, and \textit{TEMPO} improves both planners on both tasks at every goal distance. Details on the tasks and datasets can be found in Appendix~\ref{app:envs}.

\begin{table}[h]
\centering\footnotesize
\caption{\textbf{Planning results for offline world models on two RoboCasa kitchen tasks, one to three plans from the goal.} Success rate as a fraction of episodes. Bold marks the better of each pair.}
\label{tab:robocasa}
\setlength{\tabcolsep}{4pt}
\footnotesize %
\scalebox{0.65}{%
\begin{tabular}{lcc}
\toprule
\textbf{Planner} & \textbf{Pick-and-Place} (SR $\uparrow$) & \textbf{Kitchen-Nav} (SR $\uparrow$) \\
\midrule
\rowcolor{gray!12}\multicolumn{3}{l}{\textbf{LeWM}} \\
one plan & 0.32 & 0.16 \\
\quad \textbf{+ TEMPO (ours)} & \textbf{0.36} & \textbf{0.22} \\
\addlinespace[1.5pt]
two plans & 0.20 & 0.19 \\
\quad \textbf{+ TEMPO (ours)} & \textbf{0.21} & \textbf{0.27} \\
\addlinespace[1.5pt]
three plans & 0.14 & 0.23 \\
\quad \textbf{+ TEMPO (ours)} & \textbf{0.16} & \textbf{0.26} \\
\addlinespace[4pt]
\midrule
\rowcolor{gray!12}\multicolumn{3}{l}{\textbf{PLDM}} \\
one plan & 0.27 & 0.14 \\
\quad \textbf{+ TEMPO (ours)} & \textbf{0.32} & \textbf{0.26} \\
\addlinespace[1.5pt]
two plans & 0.16 & 0.18 \\
\quad \textbf{+ TEMPO (ours)} & \textbf{0.18} & \textbf{0.22} \\
\addlinespace[1.5pt]
three plans & 0.10 & 0.20 \\
\quad \textbf{+ TEMPO (ours)} & \textbf{0.12} & \textbf{0.24} \\
\bottomrule
\end{tabular}}
\end{table}

\section{Analysis of the temporal distance}
\label{sec:why}
\paragraph{Goal distance measured in $k$ plans.} Every planner optimises one short plan at a time: $H = 5$ action blocks of five environment
steps, 25 steps in all. The CEM search picks the best 25-step action sequence, executes it in full, re-encodes the new observation and
plans again (receding-horizon MPC). A goal one plan away is the frame 25 steps later in a recorded trajectory, so a single plan can
reach it. A goal three plans away is 75 steps out, so no single plan reaches it and the planner must chain at least three plans,
replanning after each, within its step budget. Nothing about the planner changes across goal distances; only the distance to the goal
does.

\paragraph{Why the flat planner stalls at range.} The planner never scores the goal itself. In each round it can only choose the plan
whose imagined end state, 25 steps ahead, scores best under its cost. The published cost asks of every candidate, ``how similar does
your end state look to the goal frame?'' It behaves much like a straight-line distance: it prefers end states that point directly at the
goal. One plan from the goal, that question and ``how close are you to finishing?'' have the same answer. Three plans away they do not.
On TwoRoom the path must bend through the door, and in a maze it bends around walls, so the shortest route first moves away from the
goal. The latent distance scores that detour as worse, and the imagined state most similar to the goal frame is the one pressed against
the wall on the goal's side, dozens of steps from the goal through the door; the planner drives into the wall and stalls
(Appendix~\ref{sec:geometry}). Equivalently, the latent distance defines a convex basin around the goal, whereas the task is not convex: walls split
the basin, and a planner that descends it greedily is trapped in a local minimum.

\paragraph{What \textit{TEMPO} changes.} $\lVert F_\phi(\hat z_{t+H}) - F_\phi(G)\rVert$ asks instead ``how many steps are you from the
goal?'' $F_\phi$ is trained on real trajectories to predict how many steps separate two states, so its distance bends around walls:
trajectories that reached the other room went through the door, so a state next to the door, even one farther from the goal in a
straight line, receives a small step count, and a state pressed against the wall on the wrong side receives a large one. The result
behaves like a shortest-path (geodesic) distance, monotone along the path in a way raw latent similarity is not. Descending it leads
through the door rather than into the wall; each 25-step plan now picks an end state that makes real progress, and three such plans
chain to the goal. Everything else stays
as published, the frozen model, the CEM search, the 25-step plans executed in full and replanned in closed loop, so the gain over the
flat planner isolates the effect of the objective alone.

\paragraph{$F_\phi$ is a distance, not a clock.} Since $F_\phi$ is trained on pairs from the same episode, it could in principle
learn the frame index, $F_\phi(z_t) \approx t$. Table~\ref{tab:fclock} rules this out. States with the same frame index but from
different episodes lie farther apart than states 25 steps apart in one episode, the frame index is no more decodable from $F_\phi$
than from the raw latent, and in the navigation environments the $F_\phi$-distance between episodes tracks the agent's position
(Spearman $\rho = 0.93$ on TwoRoom, $0.67$ on Wall).

\begin{table}[h]
\centering\footnotesize
\caption{$F_\phi$ does not encode the frame index. Median $F_\phi$-distance (units of 25 steps) between states with the same
frame index in different episodes, against states 25 steps apart in the same episode, and $R^2$ of a linear probe predicting the
frame index from $F_\phi(z)$ and from the raw latent $z$.}
\label{tab:fclock}
\setlength{\tabcolsep}{4pt}
\footnotesize
\scalebox{0.75}{%
\begin{tabular}{lcccc}
\toprule
& \multicolumn{2}{c}{\textbf{$F_\phi$-distance}} & \multicolumn{2}{c}{\textbf{$R^2$ of frame index}} \\
\cmidrule(lr){2-3}\cmidrule(lr){4-5}
\textbf{Environment} & \textbf{same index, other episode} & \textbf{25 steps, same episode} & \textbf{from $F_\phi(z)$} & \textbf{from $z$} \\
\midrule
TwoRoom & 1.48 & 1.03 & 0.11 & 0.13 \\
Push-T & 2.27 & 1.09 & 0.54 & 0.73 \\
Reacher & 2.27 & 1.95 & 0.01 & 0.03 \\
Cube & 1.63 & 1.07 & 0.03 & 0.13 \\
Scene & 2.45 & 1.03 & 0.21 & 0.32 \\
Wall & 1.09 & 0.70 & 0.07 & 0.10 \\
\bottomrule
\end{tabular}}
\end{table}

\paragraph{$F_\phi$ is goal-agnostic and generalises within an environment.} $F_\phi$ is trained per environment on the world model's
latent, as the world model itself is, but its loss uses no goal, reward or success label. It learns the number of steps between
arbitrary pairs of states of the same episode, and the goal is supplied only at test time, so it encodes the environment's temporal
structure rather than which states are close to a particular goal. The same recipe is used unchanged across all environments. Within
an environment $F_\phi$ also generalises beyond its training data. On PointMaze medium and large and on Ant-U, where the world model and
$F_\phi$ are trained on 90\% of the episodes, it measures step counts as accurately on the held-out 10\% as on training episodes
(Table~\ref{tab:foverfit}), and without retraining it transfers to unseen configurations (Table~\ref{tab:gen}) and to start and goal
states from different episodes (cross-room, Table~\ref{tab:complex}).

\begin{table}[h]
\centering\footnotesize
\caption{\textbf{$F_\phi$ generalises to held-out episodes.} Spearman $\rho$ between the $F_\phi$-distance and the step gap $k$, and
mean absolute error of the $F_\phi$-distance against its target $k/25$ (units of 25 steps), over about 24{,}000 same-episode pairs
with $k \le 75$ (LeWM, mean over three seeds).}
\label{tab:foverfit}
\setlength{\tabcolsep}{4pt}
\footnotesize
\scalebox{0.75}{%
\begin{tabular}{lcccc}
\toprule
& \multicolumn{2}{c}{\textbf{Training episodes}} & \multicolumn{2}{c}{\textbf{Held-out episodes}} \\
\cmidrule(lr){2-3}\cmidrule(lr){4-5}
\textbf{Environment} & $\rho$ & \textbf{error} & $\rho$ & \textbf{error} \\
\midrule
PointMaze M & 0.83 & 0.28 & 0.83 & 0.28 \\
PointMaze L & 0.99 & 0.06 & 0.99 & 0.06 \\
Ant-U & 0.56 & 0.56 & 0.51 & 0.59 \\
\bottomrule
\end{tabular}}
\end{table}

\section{Quantifying the collapse of the latent distance}
\label{sec:geometry}

Executing many candidate plans from the same state and recording which reach the goal, we find that three plans from the goal the
latent distance predicts where a plan ends but not whether it succeeds: its best-scored plans succeed far less often than the best ones in
hindsight, whereas one plan away the same cost works. The latent distance is a good local cost and a poor global one, and the temporal
distance supplies the coordinate it lacks at range. In this section we measure why the flat planner fails at range, by asking what its
cost actually ranks. A failure can only be
diagnosed where it happens, so we study the two environments where the flat LeWM planner collapses between one and three plans from
the goal (Table~\ref{tab:std75}: TwoRoom from $88$ to $36\%$, Push-T from $94$ to $22\%$), and use Reacher, where it does not collapse
($95$ to $90\%$), as the control. Cube barely drops ($75$ to $69\%$) and would only repeat the control, and Wall has no three-plan setting.
The one-plan rows show the same cost close to the goal, where it still works. For 60 windows per cell we
executed 50 CEM candidates from the same state and recorded the predicted latent distance, the realised end distance and eventual
success (Table~\ref{tab:actionrank}; protocol below).

\paragraph{Three plans away the cost ranks where a plan ends, not whether it succeeds.} At offset 75 the planner's latent distance
correlates with the realised end distance at $\rho = 0.66$ (Push-T), $0.66$ (TwoRoom) and $0.85$ (Reacher), and with eventual success
at $+0.02$, $-0.07$ and $+0.22$: the predictor knows where a plan goes, and on the two collapse environments that says nothing about
which plan leads to the goal. The five best-scored plans succeed in $29\%$ of Push-T windows against $65\%$ for the hindsight best five,
and in $41$ against $75\%$ on TwoRoom. Among pairs of executed candidates with different outcomes, the one predicted closer to the target
is the one that fails in $46\%$ of pairs on Push-T and $55\%$ on TwoRoom, no better than a coin.

\paragraph{One plan away the same cost works.} At offset 25 the correlation of the latent distance with success is $+0.32$ on Push-T
and $+0.21$ on TwoRoom, and the five best-scored plans succeed in $89$ and $84\%$ of windows against $100$ and $91\%$ for the hindsight
best. The latent distance is a good local cost and a poor global one. The learned temporal distance supplies the coordinate the latent
lacks at range, the number of steps a plan's end state still is from the goal: it turns $36$ into $99$ on TwoRoom, and it stops exactly
where that number is not the task, on Push-T.
\label{app:diag}

\paragraph{Action ranking (Table~\ref{tab:actionrank}).} For 60 windows per cell (20 per seed), 50 candidate action sequences of the
CEM (the final population) are each executed from the same state for 25 steps and then followed by the flat pursuit of the goal, scored
by the planner's own latent distance. Correlations are Spearman with average ranks over the sequences of a window (a binary outcome has
many ties), pooled over windows. Among success-discordant pairs of sequences (one succeeds, one fails) the sequence predicted closer to
the target is the failing one in $46.2\%$ of $13{,}505$ pairs on Push-T, $24.3\%$ of $3{,}784$ on Reacher and $55.0\%$ of $23{,}068$ on
TwoRoom (offset 75).

\begin{table}[t]
\centering\small
\caption{\textbf{What the latent distance ranks.} We diagnose the two tasks where the flat planner collapses three plans from the goal (TwoRoom from $88$ to $36\%$, Push-T from $94$ to $22\%$); Reacher, where it does not collapse, is the control, and the one-plan rows show the same cost working close to the goal. 50 executed action sequences per window, scored by the planner's own latent distance to the target; $\rho$(score, realised distance) and $\rho$(score, eventual success) over the sequences of a window; top-5 = success of the five best-scored sequences, oracle = best five in hindsight.}
\label{tab:actionrank}
\setlength{\tabcolsep}{4pt}
\footnotesize %
\scalebox{0.75}{%
\begin{tabular}{lcccccc}
\toprule
\textbf{benchmark / offset} & \textbf{windows} & \textbf{window success} & \textbf{$\rho$(score, realised dist)} & \textbf{$\rho$(score, success)} & \textbf{top-5 success} & \textbf{oracle top-5} \\
\midrule
PushT / 25 & 59 & 0.74 & $+0.83$ & $+0.32$ & 0.89 & 1.00 \\
TwoRoom / 25 & 48 & 0.80 & $+0.49$ & $+0.21$ & 0.84 & 0.91 \\
PushT / 75 & 60 & 0.28 & $+0.66$ & $+0.02$ & 0.29 & 0.65 \\
Reacher / 75 & 60 & 0.95 & $+0.85$ & $+0.22$ & 0.97 & 0.99 \\
TwoRoom / 75 & 60 & 0.42 & $+0.66$ & $-0.07$ & 0.41 & 0.75 \\
\bottomrule
\end{tabular}}
\end{table}

\paragraph{The same plans scored by \textit{TEMPO} (Table~\ref{tab:rank-tempo}).} Three plans from the goal, on all five
environments, we execute 50 candidate plans per window from the same state and score each with both costs. Scored by \textit{TEMPO}'s
cost, the ranking tracks eventual success more closely on TwoRoom ($\rho$ from $-0.08$ to $+0.17$, the five best-scored plans
succeeding in $35$ against $26\%$), Push-T ($+0.06$ to $+0.13$) and Reacher ($+0.25$ to $+0.33$, $98$ against $95\%$), and remains
competitive on Cube and Scene.

\begin{table}[h]
\centering\small
\caption{\textbf{Which executed plans each cost ranks first, three plans from the goal.} For 20 windows per seed, 50 candidate plans are executed from the same state for 25 steps and then finished by the flat planner; each candidate is scored by LeWM's latent distance and by \textit{TEMPO}'s cost. $\rho$: Spearman correlation between the (negated) cost and eventual success; top-5: success rate (\%) of the five best-scored candidates; best five: the success rate of the five candidates that actually succeed most. Mean over three seeds; bold: the better of the pair.}
\label{tab:rank-tempo}
\footnotesize %
\scalebox{0.75}{%
\begin{tabular}{lccccc}
\toprule
& \multicolumn{2}{c}{$\rho$(cost, success)} & \multicolumn{2}{c}{top-5 success} & \\
\cmidrule(lr){2-3}\cmidrule(lr){4-5}
\textbf{Environment} & latent & \textit{TEMPO} & latent & \textit{TEMPO} & best five \\
\midrule
TwoRoom & -0.08 & \textbf{+0.17} & 26 & \textbf{35} & 57 \\
Push-T & +0.06 & \textbf{+0.13} & 30 & 30 & 51 \\
Reacher & +0.25 & \textbf{+0.33} & 95 & \textbf{98} & 99 \\
Cube & +0.39 & +0.38 & 32 & 32 & 42 \\
Scene & +0.02 & \textbf{+0.04} & 18 & \textbf{21} & 60 \\
\bottomrule
\end{tabular}}
\end{table}

\end{document}